\documentclass[10pt,journal,compsoc]{IEEEtran}
\pdfoutput=1
\usepackage{times}
\usepackage{epsfig}
\usepackage{graphicx}
\usepackage{amsmath}
\usepackage{amsthm}
\usepackage{amssymb}
\usepackage{dsfont}
\usepackage{algorithm,algpseudocode}
\usepackage{color}
\usepackage{soul}
\usepackage{caption}
\usepackage{subcaption}

\usepackage{siunitx}
\usepackage{booktabs}

\usepackage[table]{xcolor}

\ifCLASSOPTIONcompsoc
  \usepackage[nocompress]{cite}
\else
  \usepackage{cite}
\fi

\usepackage[pagebackref=false,breaklinks=true,letterpaper=true,colorlinks,bookmarks=false]{hyperref}
\newif\ifdraft
\draftfalse

\definecolor{orange}{rgb}{1,0.5,0}
\definecolor{violet}{RGB}{70,0,170}
\definecolor{magenta}{RGB}{170,0,170}
\definecolor{dgreen}{RGB}{0,150,0}

\usepackage[normalem]{ulem}

\ifdraft
 \newcommand{\PF}[1]{{\color{red}{\bf PF: #1}}}
 
 \newcommand{\WL}[1]{{\color{blue}{\bf WL: #1}}}
 
  \newcommand{\MS}[1]{{\color{dgreen}{\bf MS: #1}}}
 
\else
 \newcommand{\PF}[1]{}
 
 \newcommand{\WL}[1]{}
 
 \newcommand{\MS}[1]{}
  
\fi

\newcommand{\comment}[1]{}
\newcommand{\parag}[1]{\vspace{0.5mm}\noindent {\bf #1}}

\newcommand{\cA}{\mathcal{A}}
\newcommand{\cD}{\mathcal{D}}
\newcommand{\cF}{\mathcal{F}}

\newcommand{\cU}{\mathcal{U}}

\newcommand{\bI}{\mathbf{I}}
\newcommand{\bH}{\mathbf{H}}

\newcommand{\bO}{\mathbf{O}}

\usepackage{multirow}

\begin{document}

\title{Counting People by Estimating People Flows}

\author{Weizhe~Liu,
			Mathieu~Salzmann,
			Pascal~Fua,~\IEEEmembership{Fellow,~IEEE}%
	\IEEEcompsocitemizethanks{
		\IEEEcompsocthanksitem W. Liu, M. Salzmann and P. Fua are with the Computer Vision Laboratory, \'{E}cole Polytechnique F\'{e}d\'{e}rale de Lausanne, Switzerland. E-mail: \{weizhe.liu, mathieu.salzmann, pascal.fua\}@epfl.ch. 
		
}
}

\markboth{IEEE TRANSACTIONS ON PATTERN ANALYSIS AND MACHINE INTELLIGENCEs}%
		{Shell \MakeLowercase{\textit{et al.}}: Bare Demo of IEEEtran.cls for Computer Society Journals}

\IEEEtitleabstractindextext{

\begin{abstract}

Modern methods for counting people in crowded scenes rely on deep networks to estimate people densities in individual images. As such, only very few take advantage of temporal consistency in video sequences, and those that do only impose weak smoothness constraints across consecutive frames.
In this paper, we advocate estimating people flows across image locations between consecutive images and inferring the people densities from these flows instead of directly regressing them. This enables us to impose much stronger constraints encoding the conservation of the number of people. As a result, it significantly boosts performance without requiring a more complex architecture. Furthermore, it allows us to exploit the correlation between people flow and optical flow to further improve the results.
We also show that leveraging people conservation constraints in both a spatial and temporal manner makes it possible to train a deep crowd counting model in an active learning setting with much fewer annotations. This significantly reduces the annotation cost while still leading to similar performance to the full supervision case. 

\end{abstract}

}

\maketitle

\IEEEdisplaynontitleabstractindextext

\IEEEpeerreviewmaketitle


\section{Introduction}\label{sec:introduction}

\IEEEPARstart{C}{rowd} counting is important for applications such as video surveillance and traffic control. For example during the current COVID-19 pandemic, it has a role to play in monitoring social distancing and slowing down the spread of the disease. Most state-of-the-art approaches rely on regressors to estimate the local crowd density in individual images, which they then proceed to integrate over portions of the images to produce people counts. The regressors typically use Random Forests~\cite{Lempitsky10}, Gaussian Processes~\cite{Chan09}, or more recently  Deep Nets~\cite{Zhang16c,Sam17,Xiong17,Sindagi17,Shen18,Liu18b,Li18f,Sam18,Shi18,Zhang20,Liu18c,Liu19g,Sam20,Idrees18,Ranjan18}. 

When video sequences are available, some algorithms use temporal consistency to impose weak constraints on successive density estimates. One way is to use an LSTM to model the evolution of people densities from one frame to the next~\cite{Xiong17}. However, this does not explicitly enforce the fact that people numbers must be strictly conserved as they move about, except at very specific locations where they can move in or out of the field of view. Modeling this was attempted in~\cite{Liu19b} but, because expressing this constraint in terms of people densities is difficult, the constraints actually enforced were much weaker. 

\begin{figure*}
\centering
\begin{tabular}{cccc}
\includegraphics[width=.23\linewidth]{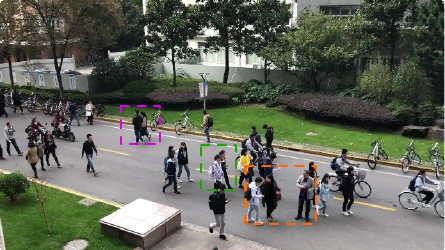}&
\includegraphics[width=.23\linewidth]{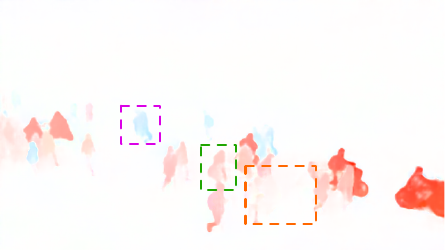}&
\includegraphics[width=.23\linewidth]{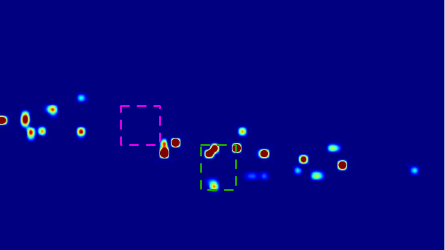}&
\includegraphics[width=.23\linewidth]{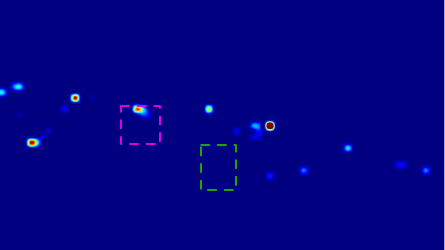}
\\
\footnotesize{(a)}&
\footnotesize{(b)}&
\footnotesize{(c)}&
\footnotesize{(d)}\\
\includegraphics[width=.23\linewidth]{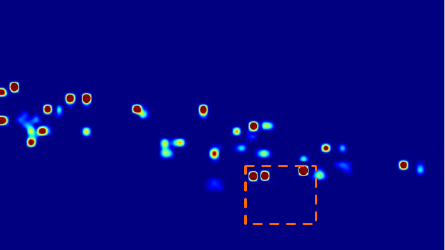}&
\includegraphics[width=.23\linewidth]{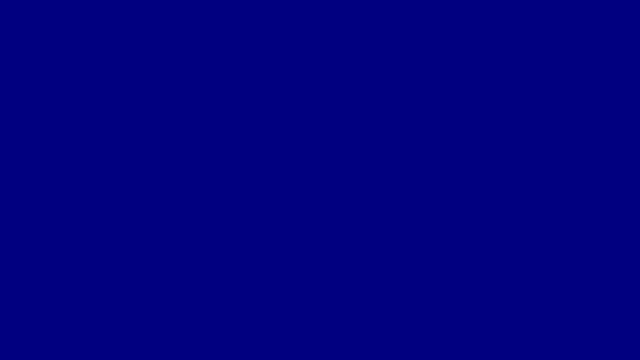}&
\includegraphics[width=.23\linewidth]{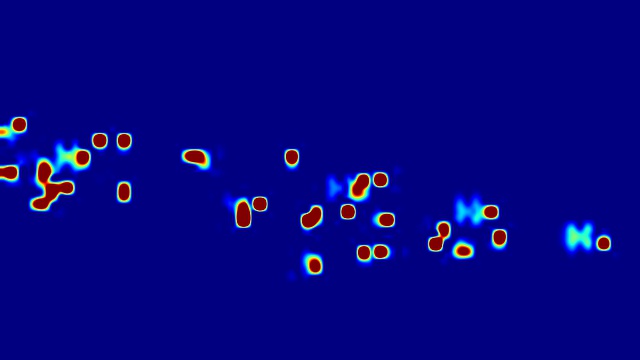}&
\includegraphics[width=.23\linewidth]{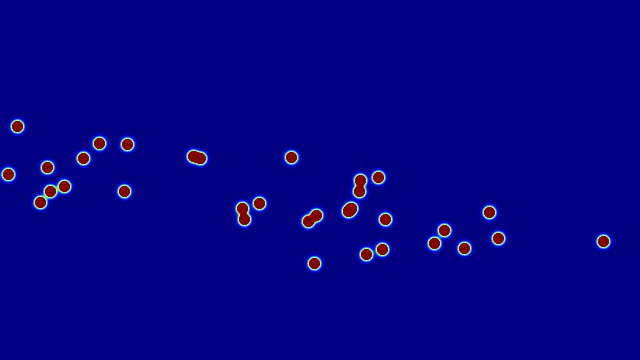}
\\
\footnotesize{(e)}&
\footnotesize{(f)}&
\footnotesize{(g)}&
\footnotesize{(h)}
\end{tabular}
\vspace{-4mm}
  \caption{ {\bf From people flow to crowd density.}  (a) Original image. (b)  Optical flow. Red denotes people moving right and blue moving left. The overlaid orange box encloses people moving slowly or not at all, the pink box people moving left, and the green box people moving right. (c) Estimated flow of people moving right. People moving left, such as those in the pink box, do not contribute to it, whereas those in the green box do. (d) Flow of people moving left. The situations within the pink and green box are reversed. (e) Estimated flow of people staying within the same grid location from one time instant to the next, such as those within the orange box.  They are not necessarily static. They may simply not have had time to change location between the two time instants. (f)~Estimated flow of people moving up. As no one does, it is almost zero everywhere. (g)~Density map inferred by summing all the flows incident on a particular location. (h)~Ground truth density map.}
  \label{fig:intro}
  \end{figure*}

In this paper, we propose to regress people flows, that is, the number of people moving from one location to another in the image plane, instead of densities. To this end, we partition the image into a number of grid locations and, for each one, we define ten potential flows, one towards each neighboring location, one towards the location {\it itself}, and the last towards regions outside the image plane. The flow towards the location itself enables us to account for people who stay in the same location from one instant to the next and the final flow to account for people who enter or exit the field of view. In our experiments, we only use it at the boundaries of the image plane because there are no occluded regions in our datasets. However, if there were occluded regions within the scene, we could simply also use that last channel for motions in and out of those. In this scenario, the places where the tenth channel is to be used would have to be scene-specific and our approach offers the required flexibility. Fig.~\ref{fig:intro} depicts some of the ten flows we compute.  All the flows incident on a grid location are summed to yield an estimate of the people density in that location. The network can therefore be trained given ground-truth estimates only of the local people densities as opposed to people flows. In other words, even though we compute flows, our network only requires ground-truth density data for training purposes, like most others.

Our formulation allows us to effectively impose people conservation constraints---people do not teleport from one region of the image to another---much more effectively than earlier approaches. This increases performance using network architectures that are neither deeper nor more complex than state-of-the-art ones. Furthermore, regressing people flows instead of densities provides a scene description that includes the motion direction and magnitude, both of which are useful for crowd analytics. This also enables us to exploit the fact that people flow and optical flow should be highly correlated, as illustrated by Fig.~\ref{fig:intro}, which provides an additional regularization constraint on the predicted flows and further enhances performance. We will demonstrate on five benchmark datasets that our approach to enforcing temporal consistency brings a substantial performance boost compared to state-of-the-art approaches. We will also show that when the cameras can be calibrated, we can apply our approach in the ground plane instead of the image plane, which further improves performance.

Another key strength of our flow-based approach is that we can use it to recast our fully-supervised approach, as described above, in an Active Learning (AL) context  that drastically reduces the supervision requirements without giving up accuracy. More specifically, our network learns to enforce the people conservation as best it can but they can still be violated. Our AL approach therefore involves first annotating a fraction of the training images,  using them to train the network, running it on the others, selecting the areas where the constraints are most violated for further human annotation, and iterating.
In effect, we use people conservation constraints to provide self-supervision and to make active learning possible. We will show that, by the time we have annotated about 6.25\% of the images, we achieve almost the same accuracy as when annotating all of them and outperform several state-of-the-art approaches trained using full supervision. 


Our contribution is therefore a novel flow-based approach to estimating people densities from video sequences that enforces strong temporal consistency constraints without requiring complex network architectures. Not only does it boost performance, it also makes it possible to implement an active-learning approach that leverages the expected consistency to reduce sixteen-fold the required amount of annotated data while preserving accuracy. 

The fully-supervised version of our framework was first introduced in a conference article~\cite{Liu20a}. We extend it here by introducing an active learning mechanism and showing that reasoning in the ground plane, instead of the image plane, further improves performance by eliminating perspective distortion artifacts.


\section{Related Work}
\label{sec:related}

Given a single image of a crowded scene, the currently dominant approach to counting people is to train a deep network to regress a people density estimate at every image location. This density is then integrated to deliver an actual count~\cite{Liu19e,Shi19a,Liu19a,Liu19b,Liu19d,Liu19f,Sindagi19a,Cheng19a,Bai20,Yang20a,liu20e,Liu20c,Zhao20a, Wang20d, Wang20e}.  In this section, we first review these approaches and then existing attempts at reducing the amount of supervision they require. 

\subsection{Regressing People Densities}

The majority of existing people counting images work on single images. We discuss them first and then move on to those that exploit temporal consistency and model people flows.

\parag{Single Image Crowd Counting.} The people density we want to measure is the number of people per unit area {\it on the ground}. However, the deep nets operate in the image plane and, as a result, the density estimate can be severely affected by the local scale of a pixel, that is, the ratio between image area and corresponding ground area. There are many ways to address this issue. For example, the algorithms of~\cite{Zhang15c,Kang17} use geometric information to adapt the network to different scene geometries. Because this information is not always readily available, other works have focused on handling the scale implicitly within the model. In~\cite{Sindagi17}, this was done by learning to predict pre-defined density levels. By contrast, the algorithms of~\cite{Onoro16,Shen18} use image patches extracted at multiple scales as input to a multi-stream network. They then either fuse the features for final density prediction~\cite{Onoro16} or introduce an {\it ad hoc} term in the training loss function~\cite{Shen18} to enforce prediction consistency across scales. In~\cite{Zhang16c,Cao18}, multi-scale features are learned by using different receptive fields and combined to predict the density. Other works propose to handle scale-variation in an adaptive way. In~\cite{Kang18}, this is done by weighing different density maps generated from input images at various scales. ~\cite{Sam17,Sam18} train an extra classifier to assign the best receptive field for each image patch. More recently,~\cite{Liu19a} proposed to extract features at multiple scales and learn how to adaptively combine them.

Instead of explicitly handling scale variations in the image plane, other single image crowd counting algorithms improve performance by using synthetic images~\cite{Wang19a, Wang20e}, leveraging auxiliary tasks~\cite{Zhao19a,Shi19b}, encoding attention mechanisms~\cite{Liu19c,Liu19d,Zhang19b,Zhang19c}, employing a Bayesian loss function~\cite{Ma19a}, exploiting multiple views~\cite{Zhang19a} or depth information~\cite{Lian19a}. We refer the reader to the recent survey~\cite{Gao20a} for more detail.

\parag{Enforcing Temporal Consistency.} While most methods work on individual images, a few have been extended to exploit temporal consistency. Perhaps the most popular way to do so is to use an LSTM~\cite{Hochreiter97}. For example, in~\cite{Xiong17},  the ConvLSTM architecture~\cite{Shi15} is used for crowd counting purposes. It is trained to enforce consistency both in the forward and the backward direction. In~\cite{Zhang17c}, an LSTM is used in conjunction with an FCN~\cite{Long15c}  to count vehicles in video sequences. A Locality-constrained Spatial Transformer (LST) is introduced in~\cite{Fang19a}.  It takes the current density map as input and outputs density maps in the next frames. The influence of these estimates on crowd density depends on the similarity between pixel values in pairs of neighboring frames.

While effective these approaches have two main limitations. First, at training time, they can only be used to impose consistency across annotated frames and cannot take advantage of unannotated ones to provide self-supervision. Second, they do not explicitly enforce the fact that people numbers must be conserved over time, except at the edges of the field of view. The recent method of~\cite{Liu19b} addresses both these issues. However, as will be discussed in more detail in Section~\ref{sec:formalization}, because the people conservation constraints are expressed in terms of numbers of people in neighboring image areas, they are much weaker than they should be.

\parag{Introducing Flow Variables.} Imposing strong conservation constraints when tracking people has been a concern long before the advent of deep learning~\cite{Pellegrini09,Vogel11,Butt13,Liu13,Collins12,Laptev07b,Gijsberts14,Butt12,Nater11,Milan14,Andriyenko10,Berclaz11}. For example, in~\cite{Berclaz11}, people tracking is formulated as  multi-target tracking on a grid and gives rise to a linear program that can be solved efficiently using the K-Shortest Path algorithm~\cite{Suurballe74}. The key to this formulation is the use as optimization variables of people flows from one grid location to another, instead of the actual number of people in each grid location. In~\cite{Pirsiavash11}, a people conservation constraint is enforced and the global solution is found by a greedy algorithm that sequentially instantiates tracks using shortest path computations on a flow network~\cite{Zhang08a}. Such people conservation constraints have since been combined with additional ones to further boost performance. They include appearance constraints~\cite{BenShitrit11,Dickle13,BenShitrit14} to prevent identity switches,  spatio-temporal constraints to force the trajectories of different objects to be disjoint~\cite{He16b}, and higher-order constraints~\cite{Butt13,Collins12}. 

However, none of these methods rely on deep learning. These kind of flow constraints have therefore never been used in a deep crowd counting context and are designed for scenarios in which people can still be tracked individually. The recent approach of~\cite{Ren20} is a good example of this. It leverages density maps and network flow constraints to improve multiple object tracking but still relies on connecting individual people detections. In this paper, we demonstrate that this approach can also be brought to bear in a deep pipeline to handle dense crowds in which people cannot be tracked as individuals anymore.

\subsection{Moving Away from Full Supervision}

There are relatively few people-counting approaches that rely on self- or weak-supervision. We discuss them below and argue that they lack some of the key features of ours.

\parag{Semi-Supervised Crowd Counting.} In~\cite{Sam19}, an autoencoder is used to learn most of the model parameters without supervision. Only those of the last two layers are learned with full supervision, which helps when there is very little annotated data but not when there is some more. In~\cite{Loy13}, only 10\% of the annotated training images are used to pre-train a model and the algorithm relies on transfer-learning to align the feature distributions across unlabeled images with similar people counts in the remaining 90\%.  Unfortunately, this method depends crucially on the quality of the pre-training. If it is not good enough, the auto-annotation of the unlabeled images is likely to cause a performance drop. Furthermore, this approach still requires image pairs from different domains that feature the same number of people, which is hard to obtain in many real world cases. Finally, it only outputs the final crowd count without a density map that denotes people's locations. Several very recent work~\cite{Sindagi20a,Liu20d} extend this auto-annotation technique by directly auto-annotating the crowd density map~\cite{Sindagi20a} or an auxiliary segmentation mask~\cite{Liu20a} based on a pre-trained model with a small amount of labeled data. As no physical world constraint is enforced in these models, the pseudo-ground truth can be very different from the true one if the labeled and unlabeled images follow different distributions.

\parag{Weakly Supervised Crowd Counting.} Another way to reduce the annotation cost is to use weak supervision, as in~\cite{Borstel16}. Instead of object-wise annotation, it relies on region-wise annotation. The image is split into arbitrarily-shaped regions that each contain two or three people. A Gaussian Process is used to map images pixels to a density map. As no localization supervision is provided, the network is prone to producing uninterpretable density maps because edges, image acquisition artifacts, and tiny fluctuations in appearance can yield larger feature changes than expected. Furthermore, manually splitting the image into regions that all contain the required number of people  is non-trivial and time consuming.

\parag{Self-Supervised Crowd Counting.} The approach of~\cite{Liu18b,Liu19g} is probably the one most related to ours.  Two extra unlabeled datasets are collected from Google by keyword searches and query-by-example image retrieval. Then, a multi-task network is trained to rank image patches  according to their crowd density, and based on the observation that any sub-image of a crowded scene image is guaranteed to contain the same number or fewer persons than the super-image. Such inequality constraints can be viewed as a weaker version of our people conservation constraints, which are equalities. However, the resulting accuracy depends on finding and properly curating the unlabeled dataset. This is a labor-intensive process because one must  ensure that the unlabeled images from the internet exhibit a similar crowd density and viewpoint angle.


\section{People Flows}


\begin{table}[t]
  \centering
  \begin{tabular}{ |l p{70mm}|}
    \hline
    $T$ &  number of time steps \\
    $K$ & number of locations in the image plane \\
    $I^{t}$ & image at $t$-th frame\\
    $m_{j}^{t}$ & number of people present at location $j$ at time $t$ \\
    $f_{i,j}^{t-1,t}$ & number of people moving from location $i$ to location $j$ between times $t - 1$ and $t$ \\
    $N(j)$ & neighborhood of location $j$ that can be reached within a single time step\\
    \hline
    \end{tabular}
  \caption{{\bf Notations.}}
  \label{tab:notations}
\end{table}

We regress {\it people flows} from images. We take these flows to be counts between two consecutive time instants of people either moving from their current location to a neighboring one, staying at the same location, or moving in or out of the field of view. They are depicted by Fig.~\ref{fig:flow} and summarized in Table~\ref{tab:notations}. People flows incident on a specific location are then summed to derive the number of people per location or {\it people count} per location. The {\it crowd density} then simply is the {\it people count} divided by the location area.  Our key insight is that this formulation enables us to impose much tighter {\it people conservation constraints} than earlier approaches. By this, we mean that we can accurately model the fact that all people present in a location at a given instant either were already there at the previous one or came from a neighboring location. This assumes the image frequency to be high enough for people not being able to move beyond neighboring locations in the time that separates consecutive frames. This is a common assumption that has proved both valid and effective in many earlier works. 

\subsection{Formalization}
\label{sec:formalization}


\begin{figure*}[htbp]
\centering
\begin{tabular}{cc}
 \includegraphics[width=.4\linewidth]{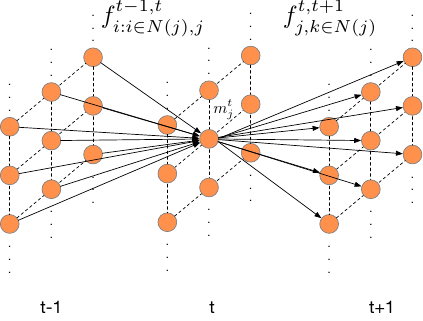}&
 \includegraphics[width=.4\linewidth]{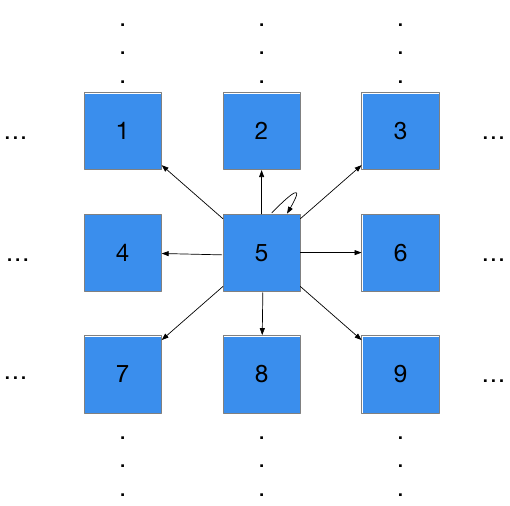}\\
 \footnotesize{(a) Grid model }&
 \footnotesize{(b) Neighborhood of each location} 
\end{tabular}
\vspace{-3mm}
  \caption{ {\bf People flows.} (a) The crowd density at time $t$ at a given location can only come from neighboring grid locations at time $t-1$ and flow to neighboring grid locations at time $t+1$, in both cases including the location itself. (b) For each location not at the boundary of the image plane, there are nine locations reachable within a single time step, including the location itself. For locations at the edge of the image plane, we add a tenth location that represents the rest of the world. It allows for flows of people who either leave the image or enter it from outside.}
    \label{fig:flow}
\end{figure*}

Let us consider a video sequence $\bI = \{\bI^1 , \ldots \bI^T\}$ and three consecutive images  $\bI^{t-1}$, $\bI^{t}$, and  $\bI^{t+1}$ from it. Let us assume that each image has been partitioned  into $K$ rectangular grid locations. In our implementation, a location is one spatial position in the final convolutional feature map, corresponding to an $8\times 8$ neighborhood in the image. However, other choices are possible.

The main constraint we want to enforce is that the number of people present at location $j$ at time $t$ is the number of people who were already there at time $t-1$ and stayed there plus the number of those who walked in from neighboring locations between $t-1$ and $t$. The number of people present at location $j$ at time $t$ also equals the sum of the number of people who stayed there until time $t+1$ and of people who went to a neighboring location between 
$t$ and $t+1$.

Let $m_j^t$ be the number of people present at location $j$ at time $t$, or {\it people count} at that location. Let $f_{i,j}^{t-1,t}$ be the number of people who move from location $i$ to location $j$ between times $t-1$ and $t$, and  $N(j)$ the neighborhood of location $j$ that can be reached within a single time step. These notations are illustrated by Fig.~\ref{fig:flow}~(a) and summarized in Table~\ref{tab:notations}. In practice, we take $N(j)$ to be the 8 neighbors of grid location $j$ plus the grid location itself to account for people who remain at the same place, as depicted by Fig.~\ref{fig:flow}~(b). Our people conservation constraint can now be written as
\begin{equation}
     \sum_{i \in N(j)} f_{i,j}^{t-1,t} = m_{j}^{t} = \sum_{k \in N(j)} f_{j,k}^{t,t+1} \;. \label{eq:flow}
\end{equation}
for all locations $j$ that are {\it not} on the edge of the grid, that is, locations from which people cannot appear or disappear without being seen elsewhere in the image. 

Most earlier approaches~\cite{Onoro16,Zhang16c,Cao18,Li18f,Liu18a,Liu19a,Liu19c} regress the values of $m_{j}^{t}$, which makes it hard to impose the constraints of Eq.~\ref{eq:flow} because many different values of the flows $f_{i,j}^{t-1,t}$ can produce the same $m_{j}^{t}$ values. For example, in~\cite{Liu19b}, the equivalent constraint is 
\begin{equation}
    \forall j \quad  m_{j}^{t}  \le \sum_{i \in N(j)} m_{i}^{t-1} \mbox{ and } m_{j}^{t}  \le \sum_{k \in N(j)}m_{k}^{t+1} \; . \label{eq:conservation}
\end{equation}
It only states that the number of people at location $j$ at time $t$ is less than or equal to the total number of people at neighboring locations at time $t-1$ and that the same holds between times $t$ and $t+1$. These are much looser constraints than the ones of Eq.~\ref{eq:flow}. They guarantee that people cannot suddenly appear but do not account for the fact that people cannot suddenly disappear either. Our formulation lets us remedy this shortcoming. By regressing the $f_{i,j}^{t-1,t}$ from pairs consecutive images and computing the values of the $m_j^t$ from these, we can impose the tighter constraints of Eq.~\ref{eq:flow}.

\subsection{Regressing the Flows}
\label{sec:regress}

We now turn to the task of training a regressor that predicts flows that correspond to what is observed while obeying the above constraints and properly handling the boundary grid locations. Let us denote the regressor that predicts the flows from $\bI^{t-1}$ and $\bI^{t}$ as $\cF$ with parameters $\Theta$ to be learned during training. In other words, $f^{t-1,t}=\cF(I^{t-1},I^{t};\Theta)$ is the vector of predicted flows between all pairs of neighboring locations between times $t-1$ and $t$. In practice, $\cF$ is implemented by a deep network. The predicted local people counts  $m_{j}^{t}$, that is, number of people per grid location $j$ and at time $t$, are taken to be the sum of the incoming flows according to Eq.~\ref{eq:flow}, and the predicted count for the whole image is the sum of all the $m_{j}^{t}$.  As the flows are not directly observable, the training data comes in the form of {\it people counts} $\bar{m}_{j}^t$ per grid location $j$ and at time $t$.

During training, our goal is therefore to find values of $\Theta$ such that 
\begin{equation}
\bar{m}_{j}^{t} = \sum_{i \in N(j)} f_{i,j}^{t-1,t} = \sum_{k \in N(j)} f_{j,k}^{t,t+1} 
\mbox{ and }
f_{i,j}^{t-1,t}  =  f_{j,i}^{t,t-1} 
\label{eq:flowConstraints}
\end{equation}
for all $i$, $j$, and $t$, except for locations at the edges of the image plane, where people can appear from and disappear to unseen parts of the scene. 

The first constraint is the people conservation constraint introduced in Section~\ref{sec:formalization}. The second accounts for the fact that,  were we to play the video sequence in reverse, the flows should have the same magnitude but the opposite direction. As will be discussed below, we enforce these constraints by incorporating them into the loss function we minimize to learn $\Theta$. Finally, we impose that all the flows be non-negative by using ReLU activations

in the network that implements $\cF$.  Note that we only require the people flows to be non-negative; the fact that a location may contain less than 1 person simply means that the flow value will be less than~1.

\begin{figure*}[htbp]
\centering
 \includegraphics[width=1.0\linewidth]{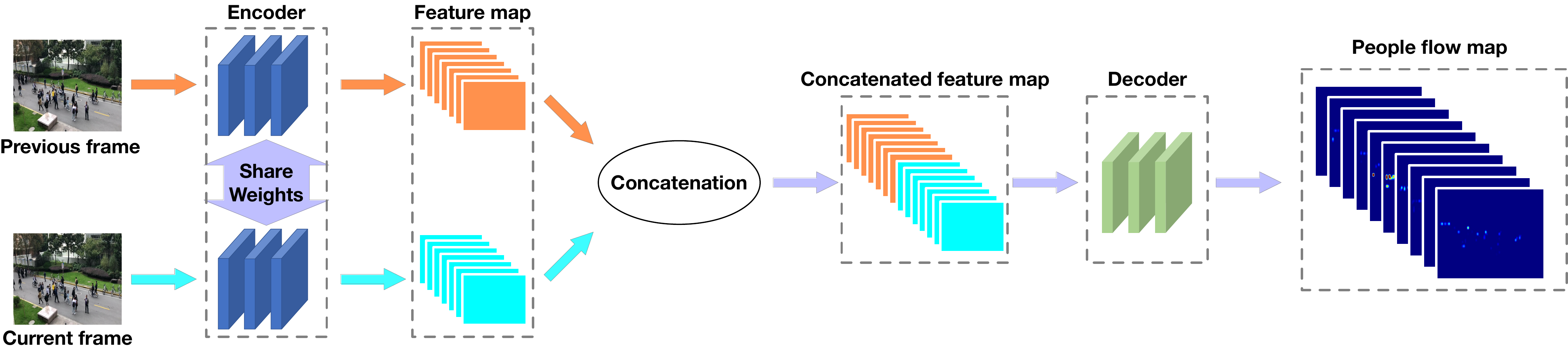}
\vspace{-3mm}
  \caption{ {\bf Model architecture:} Two consecutive RGB image frames are fed to the same encoder network that relies on the CAN scale-aware feature extractor of~\cite{Liu19a}. These multi-scale features are further concatenated and fed to a decoder network to produce the final people flow maps.}
  \label{fig:model}
\end{figure*}

\parag{Regressor Architecture.} Recall that $f^{t-1,t}=\cF(\bI^{t-1},\bI^{t};\Theta)$ is a vector of predicted flows from neighboring locations between times $t-1$ and $t$. In practice, $\cF$ is implemented by the encoding/decoding architecture shown in Fig.~\ref{fig:model}, and $f^{t-1,t}$ has the same dimension as the image grid and 10 channels per location. The first are the flows to the 9 possible neighbors depicted by Fig.~\ref{fig:flow}~(b) and the tenth represents potential flows from outside the image and is therefore only meaningful at the edges. The fifth channel denotes the flow towards the location itself, which enables us to account for people who stay in the same location from one instant to the next. 

To compute $f^{t-1,t}$, consecutive frames $\bI^{t-1}$ and $\bI^{t}$ are fed to the CAN encoder network of~\cite{Liu19a}. This yields deep features $s^{t-1} = \mathcal{E}_{e}(I^{t-1};\Theta_{e})$ and $s^{t} = \mathcal{E}_{e}(I^{t};\Theta_{e})$, where $\mathcal{E}_{e}$ denotes the encoder with weights $\Theta_{e}$. These features are then concatenated and fed to a decoder network to output $f^{t-1,t}=\mathcal{D}(s^{t-1},s^{t};\Theta_{d})$, where $\mathcal{D}$ is the decoder with weights $\Theta_{d}$. $\mathcal{D}$ comprises the back-end decoder of CAN~\cite{Liu19a} with an additional final ReLU layer to guarantee that the output is always non-negative. The encoder and decoder specifications are given in the supplementary material.

\parag{Grid Size.} In all our experiments, we treated each spatial location in the output people flow map as a separate location. Since our CAN~\cite{Liu19a} backbone outputs a down-sampled density map, each output grid location represents an 8 $\times$ 8 pixel block in the input image. This down-sampling rate is common in crowd counting models~\cite{Liu19a,Liu19b,Li18f}  because it represents a good compromise between high-resolution  of the density map and efficiency of the model. In the supplementary material, we will confirm this by showing that changing the down-sampling rate degrades performance.

\parag{Loss Function and Training.} To obtain the ground-truth maps $\bar{m}^t$ of Eq.~\ref{eq:flowConstraints}, we use the same approach as in most previous work~\cite{Onoro16,Zhang16c,Cao18,Li18f,Liu18a,Liu19a,Liu19c}. In each image $\bI^{t}$, we annotate a set of $s^{t}$ 2D points $P^{t} = {\{P^{t}_i} \}_{1 \leq i \leq s^t}$ that denote the positions of the human heads in the scene. The corresponding ground-truth density map $\bar{m}^t$ is obtained by convolving an image containing ones at these locations and zeroes elsewhere with a Gaussian kernel $\mathcal{N}( \cdot |\mu,\sigma^{2})$ with mean $\mu$ and standard deviation $\sigma$.  We write
\begin{eqnarray}
    \bar{m}^{t}_j = \sum_{i=1}^{s^t}\mathcal{N}(p_j|\mu=P^{t}_i,\sigma^{2})\;,\; \forall j\;,
    \label{eq:densityMap}
\end{eqnarray}
where $p_j$ denotes the center of location $j$.  Note that this formulation preserves the constraints of Eq.~\ref{eq:flowConstraints} because we perform the same convolution across the whole image. In other words, if a person moves in a given direction by $n$ pixels, the corresponding contribution to the density map will shift in the same direction and also by $n$ pixels.


The final ReLU layer of the regressor guarantees that the estimated flows are non-negative. To enforce the constraints of Eq.~\ref{eq:flowConstraints}, we take our combined loss function $ L_{combi} $ to be the weighted sum of two loss terms. We write
\begin{small}
\begin{align}
   L_{combi} &= \sum_t L_{flow}^t+\alpha L_{cycle}^t \; , \label{eq:loss1} \\
  L_{flow}^t &= \sum_{j \in I^{t}}\hspace{-1mm}\left[(\bar{m}_{j}^{t}  - \sum_{i \in N(j)} \hspace{-2mm} f_{i,j}^{t-1,t} )^{2} + (\bar{m}_{j}^{t}  - \sum_{k \in N(j)} \hspace{-2mm} f_{j,k}^{t,t+1} )^{2}\right] , \nonumber\\
  L_{cycle}^t & = \sum_{j \in I^{t}}\hspace{-1mm}\left[\hspace{-0mm}\sum_{i \in N(j)}\hspace{-2mm}(f_{i,j}^{t-1,t} \hspace{-2mm} - f_{j,i}^{t,t-1})^{2}  \hspace{-1mm}+ \hspace{-2mm}\sum_{k \in N(j)}\hspace{-1mm}(f_{j,k}^{t,t+1} \hspace{-2mm}- f_{k,j}^{t+1,t})^{2}\right] , \nonumber
\end{align}
\end{small}
where $\bar{m}_{j}^{t}$ is the ground-truth crowd density value, that is, the {\it people count} at time $t$ and location $j$ of Eq.~\ref{eq:densityMap} and $\alpha$ is a scalar weight we set to 1 in all our experiments. 

At training time, we systematically use three consecutive frames to evaluate $L_{combi}$ and our flow formulation requires a density map at consecutive triplet frames. A limitation of this formulation is that requires all frames to be annotated. In practice, this is not necessarily the case. In some of the examples we present in the results section, only one in 60 or 255 frames is annotated. Hence, let $\cA$ be the set of frames that are annotated and $\cU$ the set of their previous and next frames that are not and for which $\bar{m}^{t}$ is therefore unavailable. For these frames, it still holds that {\small $\sum_{i \in N(j)} f_{i,j}^{t-1,t} = \sum_{k \in N(j)} f_{j,k}^{t,t+1}$} for all $j$, even if the value of the sum is unknown. We therefore rewrite our loss function as
\begin{small}
    \begin{align}
     L_{combi} &= \sum_{t \in \cA} L_{flow}^t+\sum_{t \in \cU} L_{uflow}^t+\alpha \sum_t L_{cycle}^t \; , \label{eq:loss2} \\
     L_{uflow}^t &= \sum_{j \in I^{t}}( \sum_{i \in N(j)} \hspace{-2mm} f_{i,j}^{t-1,t} - \sum_{k \in N(j)} \hspace{-2mm} f_{j,k}^{t,t+1} )^2 \;, \nonumber
    \end{align}
\end{small}
where $L_{flow}$ and  $L_{cycle}$ are defined as in Eq.~\ref{eq:loss1}. Algorithm~\ref{alg:3sample} describes our training scheme in more detail. In the results section, we show that our algorithm can handle having only one in 255 frames annotated.


\begin{algorithm}
     \caption{Three-Frames Training Algorithm} \label{alg:3sample}
    \begin{algorithmic}[0]
    \Require Training image sequence $\{\bI^1,\ldots,\bI^T\}$ with an interval $V$ between keyframes.
    \Require Ground-truth density maps $\{\bar{m}^{V},\bar{m}^{2V}...,\bar{m}^{(T//V)V}\}$ computed by convolving the annotations according to Eq.~\ref{eq:densityMap}. 
    \Statex
    \Procedure{Train}{$\{\bI^1,..,\bI^T\}$,$\{\bar{m}^{V},..,\bar{m}^{(T//V)V}\}$ }
    \State Initialize the weights $\Theta$ of regressor network $\cF$
    \For{$\#$ of gradient iterations}
    \State Pick 3 consecutive frames $(\bI^{t-1},\bI^{t},\bI^{t+1})$, where $t$ is a multiple of $V$
    \If {V=1}
    \State Minimize $L_{combi}$ of Eq.~\ref{eq:loss1} w.r.t. $\Theta$ using Adam
    \Else 
    \State Minimize $L_{combi}$ of Eq.~\ref{eq:loss2} w.r.t. $\Theta$ using Adam
    \EndIf
    \EndFor
    \EndProcedure
    \end{algorithmic}
\end{algorithm}

\subsection{Exploiting Optical Flow}
\label{sec:flow}

When the camera is static, both the people flow discussed above and the optical flow that can be computed directly from the images stem for the motion of the people. They should therefore be highly correlated. In fact, this remains true even if the camera moves because its motion creates an apparent flow of people from one image location to another. However, there is no simple linear relationship between people flow and optical flow. To account for their correlation, we therefore introduce an additional loss function, which we define as
 \begin{align}
    L_{optical} &= \sum_{j} \delta(m_j>0)(\bO_j - \bar{o}^{t-1,t}_j)^2 \;,    \label{eq:loss_optical}\\
    {\rm where}\;\;\;\;    \bO            &= \cF_o(m^{t-1},m^{t};\Theta_{o}) \;,  \nonumber
\end{align}
$m^{t-1}$ and $m^{t}$ are density maps inferred from our predicted flows using Eq.~\ref{eq:flow}, $\bO_j$ denotes the corresponding predicted optical flow at grid location $j$ by a pre-trained regressor $\cF_o$, $\bar{o}^{t-1,t}$ is the optical flow from frames $t-1$ to $t$ computed by a state-of-the-art optical flow network~\cite{Sun18a}, and the indicator function $\delta(m_j>0)$ ensures that the correlation is only enforced where there are people. This is especially useful when the camera moves to discount the optical flows generated by the changing background. We also use CAN~\cite{Liu19a} as the optical flow regressor $\cF_o$ with 2 input channels, one for $m^{t-1}$ and the other $m^{t}$. This network is pre-trained separately on the training data and then used to train the people flow regressor.

Pre-training the regressor $\cF_o$ requires annotations for consecutive frames, that is, $V=1$ in the definition of Algorithm~\ref{alg:3sample}. When such annotations are available, we use this algorithm again but replace $L_{combi}$ by
\begin{eqnarray}
    L_{all} =L_{combi}+\beta L_{optical} \; .
    \label{eq:loss_all}
\end{eqnarray}
In all our experiments, we set $\beta$ to 0.0001 to account for the fact that the optical flow values are around 4,000 times larger than the people flow values. $\cF_o$ is also pre-trained with Adam and a learning rate of $1e-4$. During pre-training, $\cF_o$ maps the ground-truth density map pairs $\bar{m}^{t-1}$, $\bar{m}^{t}$ to the optical flow map $\bar{o}^{t-1,t}$ from frames $t-1$ to $t$ as
\begin{align}
    \bar{o}^{t-1,t}   = \sum_{j} \delta(\bar{m}_j>0) \cF_o(\bar{m}^{t-1},\bar{m}^{t};\Theta_{o}) \;. 
    \label{eq:optical_map}
\end{align}
This pre-trained network $\cF_o$ is then used as a regularization term when training our people flow model, using Eq.~\ref{eq:loss_optical} and Eq.~\ref{eq:loss_all}, where $m^{t-1}$ and $m^{t}$ are density maps obtained by summing our predicted flows. 


\section{Using Less Annotated Training Data} 
\label{sec:AL}

Recall from Section~\ref{sec:regress} that we annotate only a set of keyframes. In this section, we show that we do not even need to annotate them fully.  It is enough to only annotate small portions of them to pre-train the network and then exploit the flow constraints to iteratively select additional patches to be annotated. We will see in the results section that this active learning strategy allows us to achieve an accuracy that is close to what we get with full supervision at a much reduced annotation cost. 

\subsection{Patch Selection}

\begin{figure}
\centering
\includegraphics[width=1.0\linewidth]{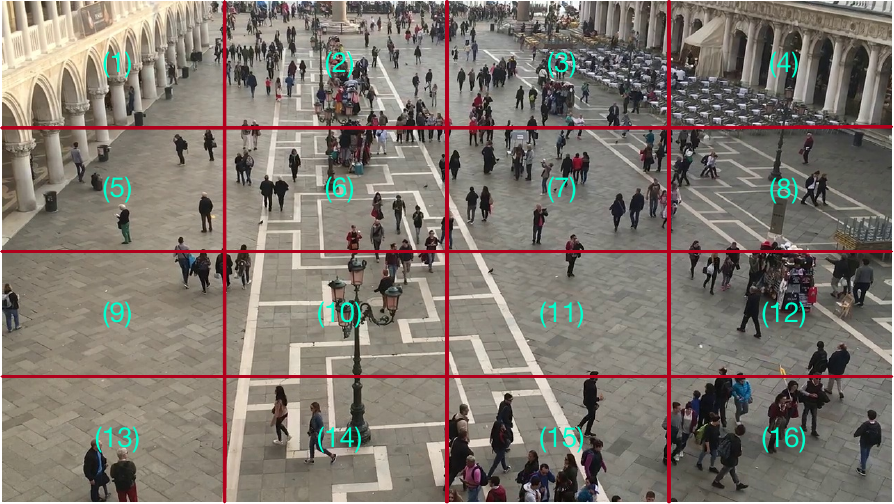}

  \caption{ {\bf Spatial people conservation constraint.}  An image from {\bf Venice}~\cite{Liu19a} dataset, we could split this image into $4 \times 4$ patches. Any adjacent $a \times a$ patches would constitute a super-patch. The spatial people conservation constraint hold between any super-patch and all the patches inside it. For example, if we only annotate the $15$th patch, one of the people conservation constraint is that the number of people in a super-patch that consists of the $11$th, $12$th, $15$th and $16$th patches, equals to the sum of the number of people in the $11$th, $12$th, $15$th and $16$th patches.} 
  \label{fig:spatial}
  \end{figure}

Let us split each keyframe image $\bI^t$ into a set of $n \times n$ patches  $P_{k}^{t}$, where $k$ is the patch index, as shown in Fig.~\ref{fig:spatial}. Instead of annotating whole images, we can annotate a single one of these patches in a subset of the keyframes and use the three-frame Algorithm~\ref{alg:3sample} to pre-train the network. Because we use relatively little training data, it is unlikely that the values of $L_{flow}$ and $L_{cycle}$ of Eq.~\ref{eq:loss1} will be zero if we evaluate the network on patches that we have {\it not} used for training purposes, at least not without further-training. In other words, the people conservation constraints of Eq.~\ref{eq:flowConstraints} will be violated. To take advantage of this, we define 
\begin{equation}
    E(P_{k}^{t}) =  \sum_{j \in P_{k}^{t} }|\sum_{i \in N(j)} f_{i,j}^{t-1,t} -\sum_{k \in N(j)} f_{j,k}^{t,t+1}| \; ,
   \label{eq:error}
\end{equation}
a measure of how much the people conversation constraint is violated within patch $P_{k}^{t}$. 

\begin{figure*}[htbp]
\centering
 \includegraphics[width=1.0\linewidth]{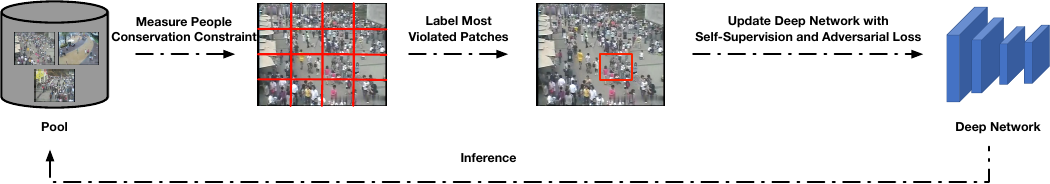}
\vspace{-3mm}
  \caption{ {\bf Our active learning pipeline.} We first annotate a fraction of the training image patches, use them to train the network while minimizing the consistency and adversarial loss terms, and then run inference on the others. We then select patches where the people conservation constraints are most violated for further human annotation and iterate the process. }
  \label{fig:al}
\end{figure*}

We then implement the simple patch selection strategy depicted by Fig.~\ref{fig:al} and detailed by Algorithm~\ref{alg:select}. In practice, we initially annotate one patch in 25\% of the keyframes, and use 60\% of them for training and the remaining 40\% for validation. We train our network by minimizing the loss function $L_{combi}$ of Eq.~\ref{eq:loss1}, whose supervised component $L_{flow}$ is only evaluated on the annotated patches. We then forward pass the remaining keyframes through our network and, within each one, annotate the patch with the larger $E$. We repeat this process 5 times, selecting 15\% of all the initially-unannotated keyframes at each such iteration and retraining the model with the newly-annotated image patches. 


\vspace{1mm}
\begin{algorithm}
     \caption{Active Patch Selection Algorithm} \label{alg:select}
    \begin{algorithmic}
    \Require $U$ unlabeled keyframes 
    \Require Pre-trained regressor network $\cF$ using $[0.25U]$ keyframes
    \Require Remaining unlabeled keyframes $\{\bI^{o1},\ldots,\bI^{o[0.75U]}\}$
    \Statex
    \Procedure{Annotation}{$\{\bI^{o1},..,\bI^{o[0.75U]}\}$ }
    \For{$\#$ of selection iterations}
    \For{$\#$ of unlabeled keyframes}
    \State Pick 3 consecutive frames $(\bI^{t-1},\bI^{t},\bI^{t+1})$, where $t$ is a multiple of $V$ (i.e., $\bI^{t}$ is a keyframe)
    \For{$\#$ of patches}
    \State Pick the $l$-th patch
    \State Compute the measure $E$ of Eq.~\ref{eq:error} 
    \EndFor
    \State Take the maximum value $E$ over all the patches in each unlabeled keyframe as the error for this keyframe 
    \EndFor
    \State Select $0.15U$ unlabeled keyframes with largest error
    \State For one, annotate the patch with highest value $E$
    \State Update the set of unlabeled keyframes
    \State Re-train $\cF$ with all the labeled keyframes
    \EndFor
    \EndProcedure
    \end{algorithmic}
\end{algorithm}

\subsection{Adding New Terms to the Objective Function}

In the fully supervised case, there was no need to enforce spatial consistency across patches in the {\it same} image because the ground-truth data did it implicitly. However, in the scenario where we have ground-truth data for only a small subset of the patches, this has to be done explicitly. Furthermore, we must avoid overfitting to the labeled patches.
  
To achieve these two goals, we introduce two additional loss terms $L_{spatial}$ and $L_{advers}$ described in the remainder of this section, and thus minimize the overall loss
\begin{eqnarray}
    L_{overall}  = L_{combi}+\gamma L_{spatial} + \delta L_{advers}  \; , 
    \label{eq:overall}
\end{eqnarray}
where $\gamma$ and $\delta$ are weighing factors. The training strategy is detailed by Algorithm~\ref{alg:self}.

\subsubsection{Spatial People Conservation Loss: $L_{spatial}$}
\label{sec:spatialLoss}


To handle the scenario where we have ground-truth data for only a subset of the patches, we replace the missing ground-truth data by spatial consistency constraints as follows. 
Let us consider keyframe $\bI^t$ that has been split into patches $\{P_{k}^{t}\}$ and assume that we have annotated $P_{j}^{t}$ only. We define $S_{k}^{t}$ as a {\it super-patch} composed of $P_{j}^{t}$ and unannotated patches $P_{k}^{t}$ for $k \in {\cal P}_j$, where ${\cal P}_j$ is a set of at most 15 indices, randomly chosen each time we compute the spatial loss. In other words, this means that a super-patch can range between the entire image and the combination of $P_{j}^{t}$ with a single immediate neighbor.  We then pass each patch through the network individually to obtain people counts $m_{k}^{t}$ for $k \in {\cal P}_j$, and further forward pass the super-patch through the network to compute people counts $M_{k}^{t}$. Because the number of people in the super-patch must be the sum of the number of people in each individual patch, we should have
\begin{equation}
\forall j, \;,  \sum_{i \in P_{j}^{t}} m_{i}^{t} + \sum_{k \in {\cal P}_j} \sum_{i \in P_{k}^{t}} m_{i}^{t} = \sum_{i \in S_{j}^{t}} M_{i}^{t}  \; . 
\label{eq:spatial}
\end{equation}
We therefore write
\begin{equation}
 L_{spatial} = \sum_j (\sum_{i \in P_{j}^{t}} m_{i}^{t} + \sum_{k \in {\cal P}_j} \sum_{i \in P_{k}^{t}} m_{i}^{t} - \sum_{i \in S_{j}^{t}} M_i^t)^2 \; .
\label{eq:spatialLoss}
\end{equation}
When we take the super-patch as input, the receptive field for the corresponding sub-patch is larger than the sub-patch itself. By contrast if we only take the sub-patch as input, the receptive field is limited to it. Therefore, our loss term encourages the estimated densities for unlabeled sub-patches to be consistent independently of the contextual information.

\subsubsection{Adversarial Loss: $L_{advers}$}

To prevent overfitting, we further introduce an adversarial loss term inspired by GANs~\cite{Goodfellow14}. We take the generator to be 
the function ${\cal G}$ that runs our flow-predicting network $\cF(\cdot,\cdot,,\Theta_{d})$ on a pair of images $(\bI^{t-1},\bI^t)$ and infers from it a people-density in $\bI^t$ by summing the flows according to Eq.~\ref{eq:flow}. We then define a  discriminator ${\cal D}(\cdot,\Theta_{d})$ as a multilayer perceptron that takes as input the people-density map $\{m_i^t , i \in P_k^t \}$ and returns the probability that it comes from a patch that has been annotated. Let $A$ be the set of patches that have been annotated. We write
\begin{eqnarray}
    L_{advers}  = - \sum_{P \in A}\log (\cD(m_{i}^{t})) - \sum_{P \notin A} \log (1-\cD(m_{i}^{t}))  \;. 
    \label{eq:adversarial}
\end{eqnarray}
%


\vspace{1mm}
\vspace{-3mm}
\begin{algorithm}
     \caption{Training with Patch Annotation} \label{alg:self}
    \begin{algorithmic}[0]
    \Require Training image sequence $\{\bI^1,\ldots,\bI^T\}$ with an interval $V$ between annotated frames.
    \Require Ground-truth density maps for one patch in every $V$ images $\{\bar{m}_{l1}^{V},\bar{m}_{l2}^{2V}...,\bar{m}_{l(T//V)}^{(T//V)V}\}$.
    \Statex
    \Procedure{Train}{$\{\bI^1,..,\bI^T\}$,$\{\bar{m}_{l1}^{V},..,\bar{m}_{l(T//V)}^{(T//V)V}\}$ }
    \State Initialize the weights $\Theta$ of regressor network $\cF$
    \For{$\#$ of gradient iterations}
    \State Pick 3 consecutive frames $(\bI^{t-1},\bI^{t},\bI^{t+1})$, where $t$ is a multiple of $V$. Only the $j$th patch of $\bI^{t}$ is annotated
    \State Reconstruct density map  $m^{t}_{j}$ using  $\cF(I_{j}^{t-1},I_{j}^{t},\Theta)$, $\cF(I_{j}^{t},I_{j}^{t+1},\Theta)$, $\cF(I_{j}^{t},I_{j}^{t-1},\Theta)$ and $\cF(I_{j}^{t+1},I_{j}^{t},\Theta)$
    \State Randomly select a patch $q$ from $(\bI^{t-1},\bI^{t},\bI^{t+1})$
    \State Reconstruct density map  $m^{t}_{q}$ using  $\cF(I_{q}^{t-1},I_{q}^{t},\Theta)$, $\cF(I_{q}^{t},I_{q}^{t+1},\Theta)$, $\cF(I_{q}^{t},I_{q}^{t-1},\Theta)$ and $\cF(I_{q}^{t+1},I_{q}^{t},\Theta)$
    \State Update $\Theta_{d}$ using $L_{advers}$ in Eq.~\ref{eq:adversarial} with RMSProp as suggested by~\cite{Arjovsky2017}
    \State Randomly select a super-patch $S_{k}^{t}$  composed of patches from  $\bI_{j}^{t}$
    \State Reconstruct density map of $S_{k}^{t}$ and other unlabeled patches inside this super-patch by passing these patches through the regressor network $\cF$ 
    \State Update $\Theta$ using $L_{overall}$ in Eq.~\ref{eq:overall} with Adam
    \EndFor
    \EndProcedure
    \end{algorithmic}
\vspace{-1mm}
\end{algorithm}

\newcommand{\oursF}[0]{{\bf OURS-FLOW}}
\newcommand{\oursC}[0]{{\bf OURS-COMBI}}
\newcommand{\oursCF}[0]{{\bf OURS-COMBI-FOR}}
\newcommand{\oursCB}[0]{{\bf OURS-COMBI-BACK}}
\newcommand{\oursO}[0]{{\bf OURS-ALL-EST}}
\newcommand{\oursI}[0]{{\bf OURS-IMG-FLOW}}
\newcommand{\oursS}[0]{{\bf OURS-COMBI-SPA}}
\newcommand{\oursG}[0]{{\bf OURS-ALL-GT}}
\newcommand{\oursGP}[0]{{\bf OURS-COMBI-GROUND}}
\newcommand{\baseline}[0]{{\bf BASELINE}}
\newcommand{\twos}[0]{{\bf IMAGE-PAIR}}
\newcommand{\ave}[0]{{\bf AVERAGE}}
\newcommand{\weak}[0]{{\bf WEAK}}

\newcommand{\patchB}[0]{{\bf PATCH-BASE}}
\newcommand{\patchBA}[0]{{\bf PATCH-BASE-AL}}
\newcommand{\patchS}[0]{{\bf PATCH-SPATIAL}}
\newcommand{\patchSA}[0]{{\bf PATCH-SPATIAL-AL}}
\newcommand{\patchA}[0]{{\bf PATCH-ALL}}
\newcommand{\patchAA}[0]{{\bf PATCH-ALL-AL}}

\newcommand{\mcnnC}[0]{{\bf MCNN-COMBI}}
\newcommand{\mcnnO}[0]{{\bf MCNN-ALL-EST}}
\newcommand{\csrC}[0]{{\bf CSRNET-COMBI}}
\newcommand{\csrO}[0]{{\bf CSRNET-ALL-EST}}

\section{Experiments}

In this section, we first introduce the evaluation metrics and benchmark datasets used in our experiments. We then show that our fully supervised approach outperforms state-of-the-art methods when operating in the image plane and does even better when image registration is available by working in the ground plane instead of the image plane. We then quantify the ability of our active learning algorithm to reduce the annotation cost. In both cases we run an ablation study to justify our choices, which we enrich in the supplementary material by studying the impact of hyper-parameters. In the supplementary material, we provide additional ablation studies about hyper-parameters, reasoning in the ground plane, settings, and variations of the proposed approach to exploit unlabeled data.

\subsection{Evaluation Metrics}
\label{sec:metrics}

Previous works in crowd density estimation use the mean absolute error ($MAE$) and the root mean squared error ($RMSE$) as evaluation metrics~\cite{Zhang16c,Zhang15c,Onoro16,Sam17,Xiong17,Sindagi17}. They are defined as 
\begin{small}
\begin{equation}
    MAE = \frac{1}{N}\sum_{i=1}^{N}|z_{i}-\hat{z_{i}}| \mbox{ and } RMSE=\sqrt{\frac{1}{N}\sum_{i=1}^{N}(z_{i}-\hat{z_{i}})^{2}} \; . \nonumber
\end{equation}
\end{small}
where $N$ is the number of test images, $z_{i}$ denotes the true number of people inside the ROI of the $i$th image and $\hat{z_{i}}$ the estimated number of people. In the benchmark datasets discussed below, the ROI is the whole image except when explicitly stated otherwise. In practice, $\hat{z_{i}}$ is taken to be $\sum_{p \in I_{i} } m_{p}$, that is, the sum over all locations or people counts obtained by summing the predicted people flows.

\subsection{Benchmark Datasets and Ground-truth Data}

\begin{figure}[htbp]
\centering
\begin{tabular}{cccc}
 \includegraphics[width=.46\linewidth]{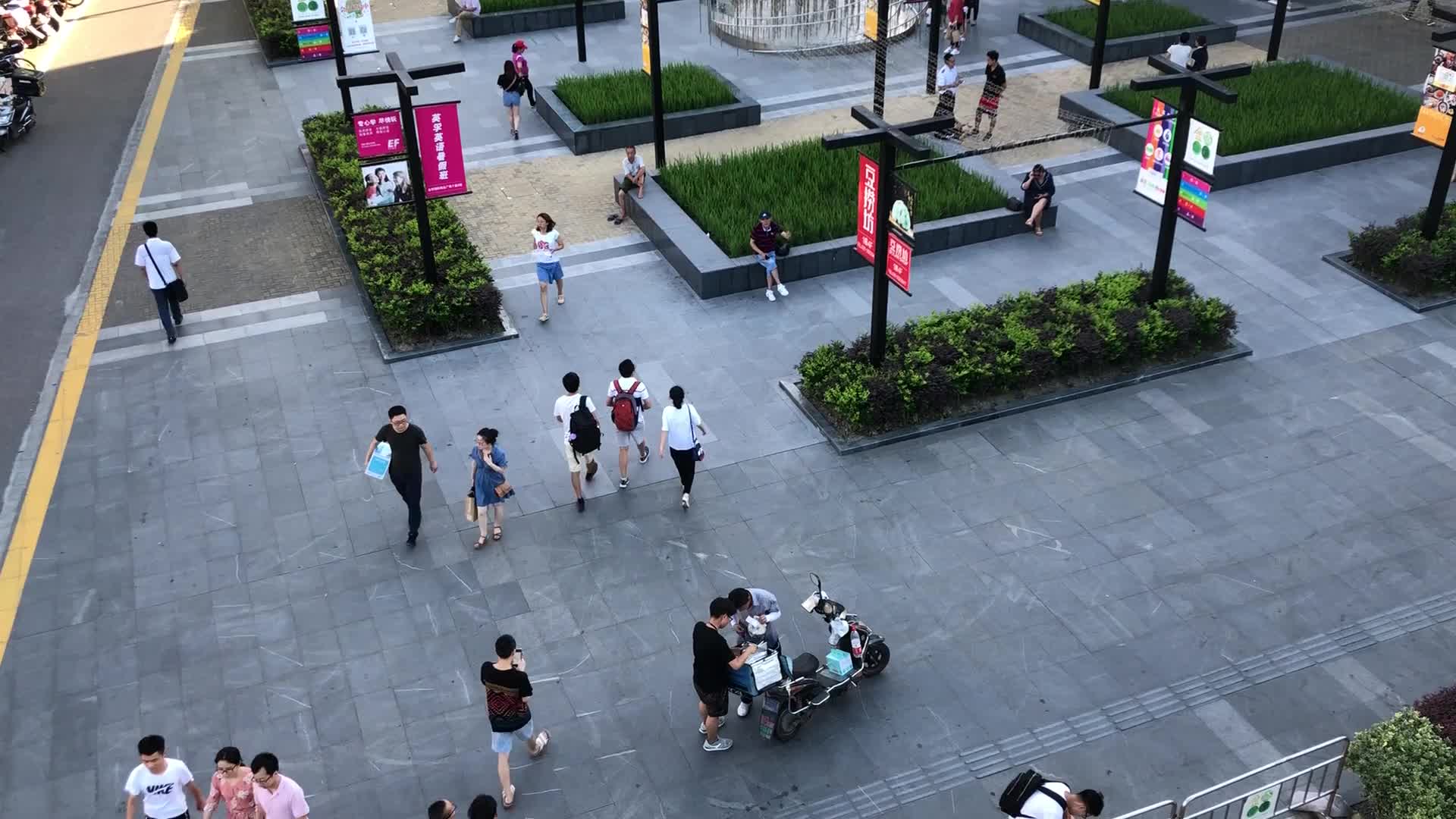}&
 \includegraphics[width=.46\linewidth]{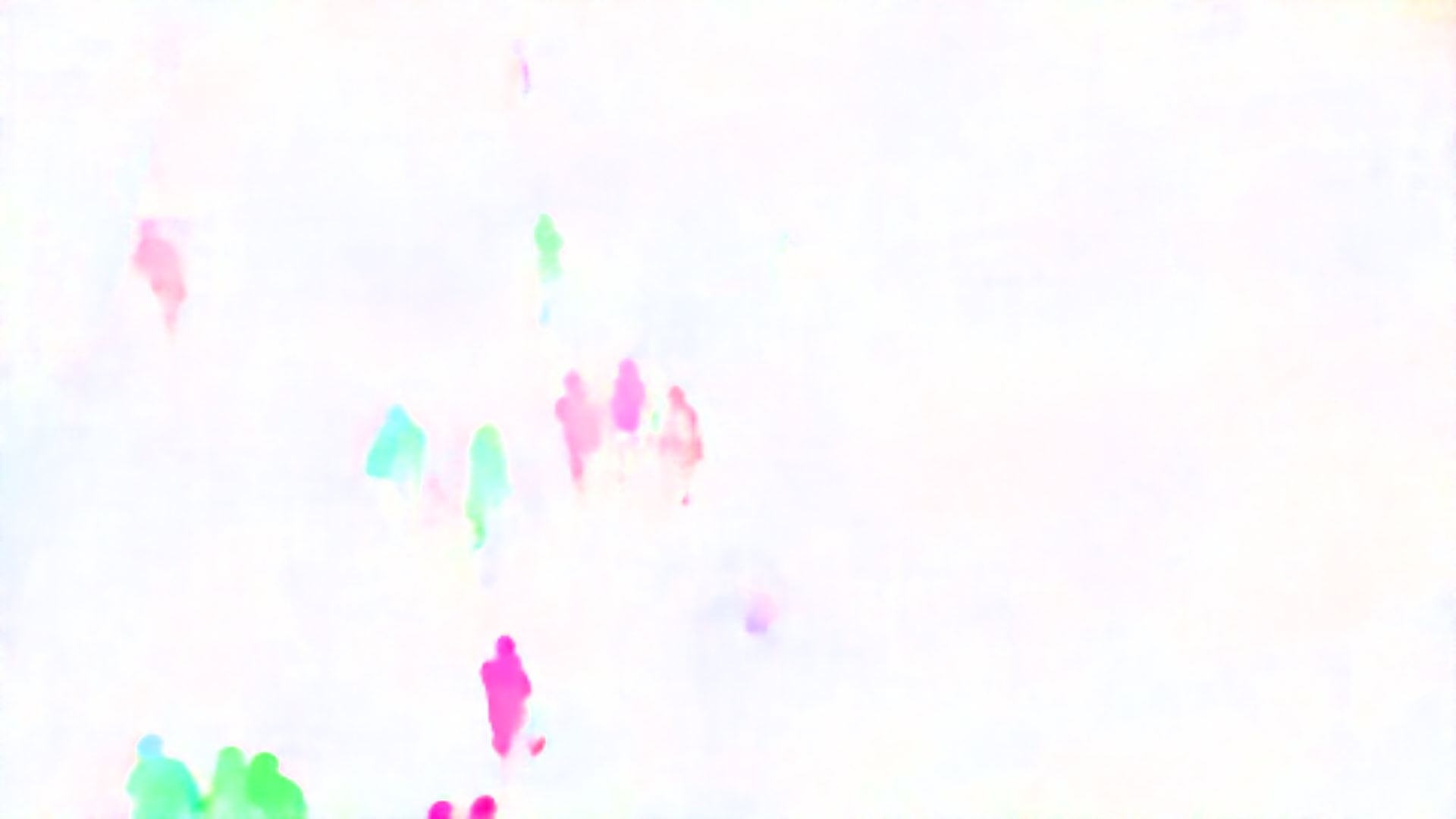}
\end{tabular}
\vspace{-3mm}
  \caption{ {\bf Estimated optical flow in  FDST.} An image and the corresponding optical flow estimated using {\bf PWC-Net}~\cite{Sun18a}.}
  \label{fig:flowFdst}
  \end{figure}

For evaluations purposes, we use five different datasets, for which the videos have been released along with recently published papers. The first one is a synthetic dataset with ground-truth optical flows. The other four are real world videos, with annotated people locations but without ground-truth optical flow. To use the optional optical flow constraints introduced in Section~\ref{sec:flow}, we therefore use the pre-trained {\bf PWC-Net}~\cite{Sun18a} to compute the loss function $L_{optical}$ of Eq.~\ref{eq:loss_optical}. Fig.~\ref{fig:flowFdst} depicts one such flow.


\begin{figure}[htbp]
\centering
\begin{tabular}{cccc}
 \includegraphics[width=.5\linewidth]{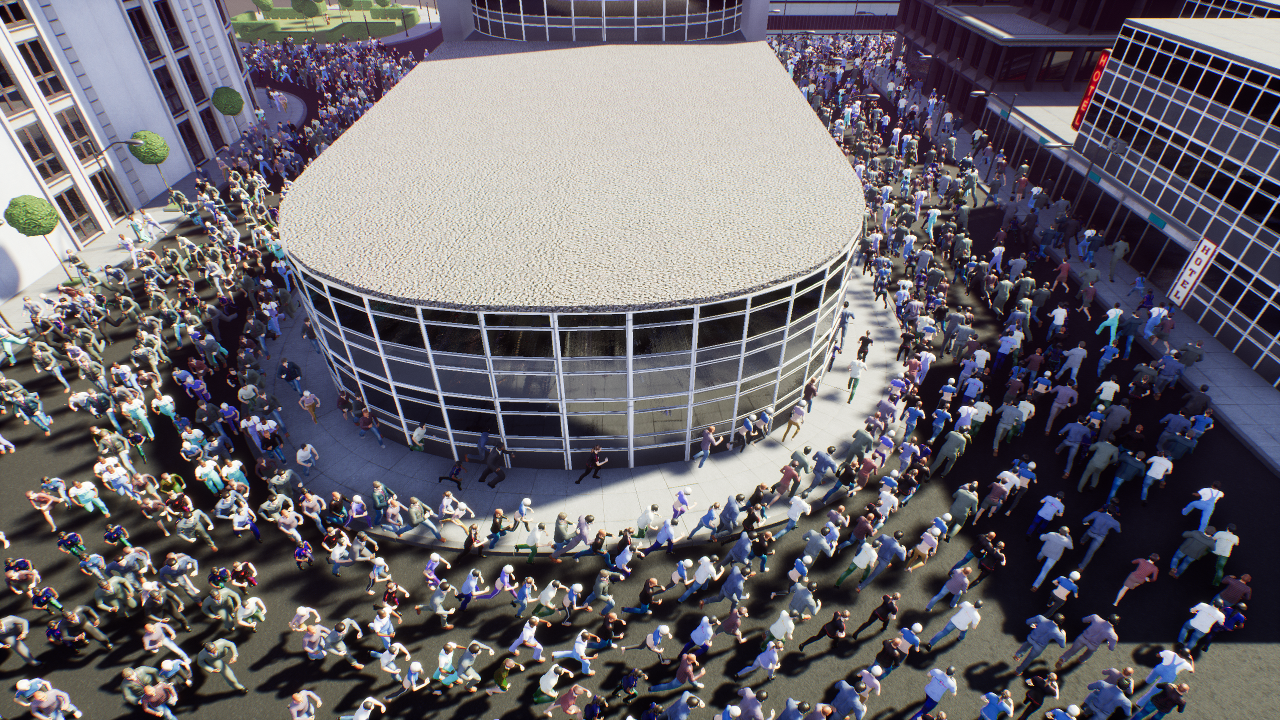}&
 \includegraphics[width=.5\linewidth]{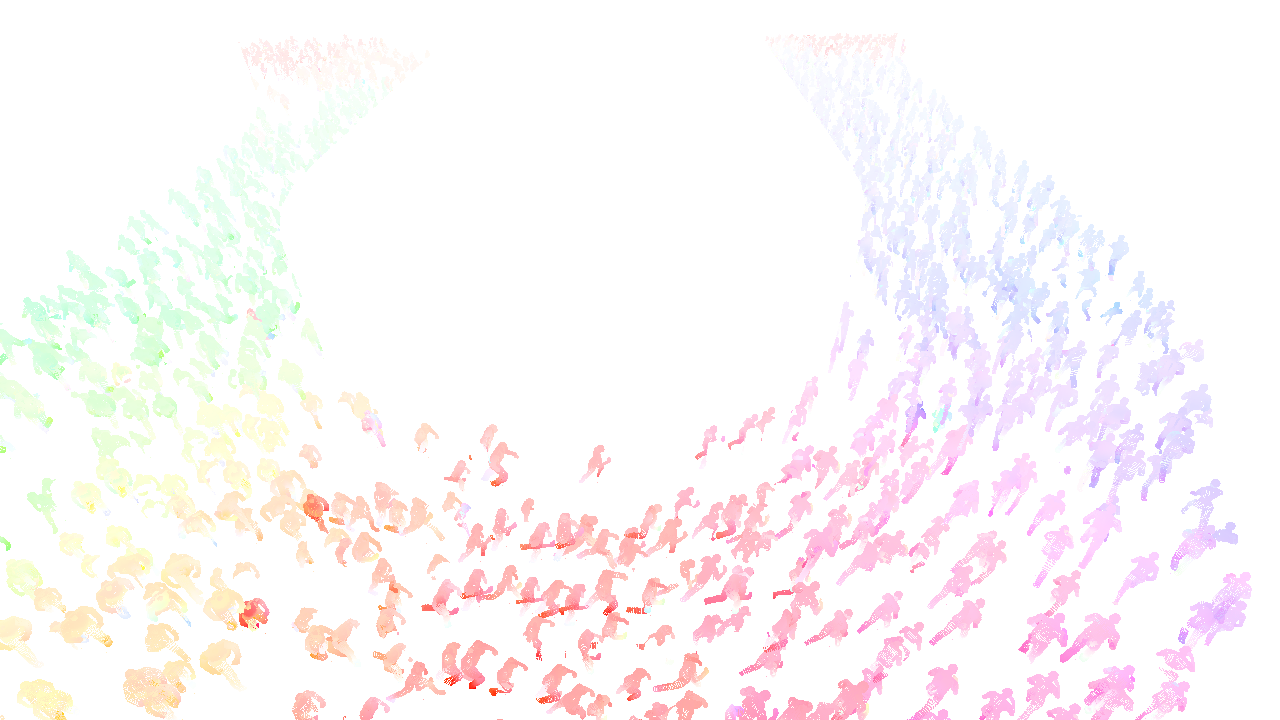}
\end{tabular}
\vspace{-3mm}
  \caption{ {\bf Ground-truth optical flow in CrowdFlow.} (left) Original image. (Right) Corresponding optical flow map.
 }
  \label{fig:flowCrowdflow}
  \end{figure}

\parag{CrowdFlow~\cite{Schroder18}.} 
This dataset consists of five synthetic sequences ranging from 300 to 450 frames each. Each one is rendered twice, once using a static camera and the other a moving one. The ground-truth optical flow is provided as shown at Fig.~\ref{fig:flowCrowdflow}. As this dataset has not been used for crowd counting before, and the training and testing sets are not clearly described in~\cite{Schroder18}, to verify the performance difference caused by using ground-truth optical flow vs. estimated one, we use the first three sequences of both the static and moving camera scenarios for training and validation, and the last two for testing.


\parag{FDST~\cite{Fang19a}.} 
It comprises 100 videos captured from 13 different scenes with a total of 150,000 frames and 394,081 annotated heads. The training set consists of 60 videos, 9000 frames and the testing set contains the remaining 40 videos, 6000 frames. We use the same setting as in~\cite{Fang19a}.

\parag{UCSD~\cite{Chan08}.} 
This dataset contains 2000 frames captured by surveillance cameras on the UCSD campus. The resolution of the frames is 238 $\times$ 158 pixels and the framerate is 10 fps.  For each frame, the number of people varies from 11 to 46. We use the same setting as in~\cite{Chan08}, with frames 601 to 1400 used as training data and the remaining 1200 frames as testing data.

\parag{Venice~\cite{Liu19a}.}
It contains 4 different sequences and in total 167 annotated frames with fixed 1,280 $\times$ 720 resolution. As in~\cite{Liu19a}, 80 images from a single long sequence are used as training data. The remaining 3 sequences are used for testing purposes.

\parag{WorldExpo'10~\cite{Zhang15c}.} 
It comprises 1,132 annotated video sequences collected from 103 different scenes. There are 3,980 annotated frames, 3,380 of which are used for training purposes. Each scene contains a Region Of Interest (ROI)  in which the people are counted. As in previous work~\cite{Zhang15c,Zhang16c,Sam17,Sam18,Li18f,Cao18,Liu18a,Sindagi17,Shen18,Ranjan18,Shi18} on this dataset, we report the \textit{MAE} of each scene, as well as the average over all scenes.

For {\bf CrowdFlow}, {\bf FDST} and {\bf UCSD}, all frames in the training set are annotated. For {\bf Venice} and {\bf WorldExpo'10}, annotations are only available for every $60$ and $255$ frames, respectively.

\subsection{Fully Supervised Approach}

\subsubsection{Comparing against Recent Techniques}
\label{sec:results}


\begin{table*}
  \begin{tabular}{ccc}
    \begin{minipage}{.3\linewidth}
  \centering
  \scalebox{0.8}{
    \begin{tabular}{lccc}
      \toprule
  Model  & Temporal&$MAE$ & $RMSE$  \\
  \midrule
  MCNN~\cite{Zhang16c} & & 172.8 & 216.0  \\
  CSRNet\cite{Li18f}& & 137.8 & 181.0\\
  CAN\cite{Liu19a} && 124.3 & 160.2 \\
  \oursC{} & \checkmark& 97.8 &  112.1 \\
  \oursO{} & \checkmark&96.3 &  111.6 \\
  \oursG{} & \checkmark &\textbf{90.9} &  \textbf{110.3} \\
  \bottomrule
  \end{tabular}
  }
\end{minipage} &

\begin{minipage}{.3\linewidth}
  \centering
  \scalebox{0.8}{
    \begin{tabular}{lccc}
      \toprule
      Model  & Temporal & $MAE$ & $RMSE$  \\
      \midrule
      MCNN~\cite{Zhang16c}&  & 3.77 & 4.88  \\
      ConvLSTM~\cite{Xiong17}& \checkmark & 4.48 & 5.82    \\
      WithoutLST~\cite{Fang19a}& & 3.87 & 5.16 \\
      LST~\cite{Fang19a} & \checkmark& 3.35 & 4.45 \\
      CAN~\cite{Liu19a} & & 2.44 & 2.96\\
      \oursC{} & \checkmark& 2.17 & 2.62  \\
      \oursO{}& \checkmark &  \textbf{2.10}  &  \textbf{2.46} \\
      \bottomrule
      \end{tabular}
      }
\end{minipage} &

\begin{minipage}{.3\linewidth}
  \centering
  \scalebox{0.8}{
    \begin{tabular}{lccc}
      \toprule
      Model & Temporal & $MAE$ & $RMSE$  \\
      \midrule
      MCNN~\cite{Zhang16c}& & 145.4 & 147.3  \\
      Switch-CNN~\cite{Sam17}& & 52.8 & 59.5    \\
      CSRNet\cite{Li18f}& & 35.8 & 50.0\\
      CAN\cite{Liu19a}& & 23.5 & 38.9 \\
      ECAN\cite{Liu19a}& & 20.5 & 29.9 \\
      GPC\cite{Liu19b} &\checkmark& 18.2 & 26.6 \\
      \oursC{} &\checkmark &  15.0  &  19.6   \\
      \oursO{} &\checkmark& 14.2  &  18.4  \\
      \oursGP{} &\checkmark &  \textbf{12.3}  &  \textbf{17.1}   \\
      \bottomrule
      \end{tabular}
      }
\end{minipage} \\

\footnotesize{(a)}&
\footnotesize{(b)} &
\footnotesize{(c)} \\
\begin{minipage}{.3\linewidth}
  \centering
  \scalebox{0.9}{
    \begin{tabular}{lccc}
      \toprule
  Model & Temporal  & MAE & RMSE  \\
  \midrule
  Zhang \textit{et al.}~\cite{Zhang15c} & & 1.60 & 3.31  \\
  Hydra-CNN~\cite{Onoro16} & & 1.07 & 1.35  \\
  CNN-Boosting~\cite{Walach16} & & 1.10 & -  \\
  MCNN~\cite{Zhang16c} & & 1.07 & 1.35  \\
  Switch-CNN~\cite{Sam17}& & 1.62 & 2.10   \\
  ConvLSTM~\cite{Xiong17}& \checkmark  & 1.30 & 1.79  \\
  Bi-ConvLSTM~\cite{Xiong17} &\checkmark & 1.13 & 1.43   \\
  ACSCP~\cite{Shen18} & & 1.04 & 1.35  \\
  CSRNet~\cite{Li18f} & & 1.16 & 1.47  \\
  SANet~\cite{Cao18} & & 1.02 & 1.29  \\
  ADCrowdNet~\cite{Liu19c}& & 0.98 & 1.25  \\
  PACNN~\cite{Shi19a} & & 0.89 & 1.18  \\
  SANet+SPANet~\cite{Cheng19a}& & 1.00 & 1.28\\
  CAN~\cite{Liu19a} & & 0.98 & 1.26\\
  \oursC{} & \checkmark&  0.86 & 1.13   \\
  \oursO{} & \checkmark &  \textbf{0.81} &  \textbf{1.07} \\
  \bottomrule
  \end{tabular}
  }
\end{minipage} &

\multicolumn{2}{c}{
  \begin{minipage}{.5\linewidth}
    \centering
    \scalebox{0.9}{
      \begin{tabular}{lcccccc|c}
        \toprule
      Model  & Temporal & Scene1 & Scene2 & Scene3 & Scene4 & Scene5 &{\bf Average} \\
      \midrule
      Zhang \textit{et al.}~\cite{Zhang15c} & & 9.8 & 14.1 & 14.3 & 22.2 & 3.7 & 12.9  \\
      MCNN~\cite{Zhang16c} & & 3.4 & 20.6 & 12.9 & 13.0 & 8.1 & 11.6 \\
      Switch-CNN~\cite{Sam17}& & 4.4 & 15.7 & 10.0 & 11.0 & 5.9 & 9.4  \\
      CP-CNN~\cite{Sindagi17} & & 2.9 & 14.7 & 10.5 & 10.4 & 5.8 & 8.9  \\
      ACSCP~\cite{Shen18} & & 2.8 & 14.05 & 9.6 & 8.1 & 2.9 & 7.5 \\
      IG-CNN~\cite{Sam18} & & 2.6 & 16.1 & 10.15 & 20.2 & 7.6 & 11.3 \\
      ic-CNN\cite{Ranjan18}& & 17.0 & 12.3 & 9.2 & 8.1 & 4.7 & 10.3\\
      D-ConvNet~\cite{Shi18}& & \textbf{1.9} & 12.1 & 20.7 & 8.3 & \textbf{2.6} & 9.1\\ 
      CSRNet~\cite{Li18f} & & 2.9 & 11.5 & 8.6 & 16.6 & 3.4 & 8.6 \\
      SANet~\cite{Cao18} & & 2.6 & 13.2 & 9.0 & 13.3 & 3.0 & 8.2 \\
      DecideNet~\cite{Liu18a}& & 2.0 & 13.14 & 8.9 & 17.4 & 4.75 & 9.23 \\
      CAN~\cite{Liu19a} & & 2.9 & 12.0 & 10.0 & \textbf{7.9} & 4.3 & 7.4  \\
      ECAN~\cite{Liu19a} & & 2.4 & \textbf{9.4} & 8.8 & 11.2  & 4.0 & 7.2  \\
      PGCNet~\cite{Yan19a}& & 2.5 & 12.7 & 8.4 & 13.7 & 3.2 & 8.1\\
      \oursC{} & \checkmark & 2.2  & 10.8  & \textbf{8.0} & 8.8 & 3.2 &  6.6  \\
      \oursO{} &\checkmark & 2.1  & 10.9 &  8.5 & 8.4 & 3.0 &  \textbf{6.58}  \\
       \bottomrule
      \end{tabular}
    }
  \end{minipage} 

} \\

\footnotesize{(d)}&
\multicolumn{2}{c}{\footnotesize{(e)}}
\end{tabular}
\caption{ {\bf Comparative results on different datasets.}  (a) {\bf CrowdFlow}. (b) {\bf FDST}. (c) {\bf Venice}. (d) {\bf UCSD}. (e) {\bf WorldExpo'10}.}
  \label{tab:eva}
\end{table*}

We denote our model trained using the combined loss function $L_{combi}$ of Section~\ref{sec:regress} as \oursC{} and the one using the full loss function $L_{all}$ of Section~\ref{sec:flow} with ground-truth optical flow as \oursG{}. In other words,  \oursG{} exploits the optical flow while \oursC{} does not. If the ground-truth optical flow is not available, we use the optical flow estimated by {\bf PWC-Net}~\cite{Sun18a} and denote this model as \oursO{}. 

\begin{figure*}[htbp]
\centering
\begin{tabular}{cccc}
\includegraphics[width=.245\linewidth]{images/crowdflow/img.png}&
 \includegraphics[width=.245\linewidth]{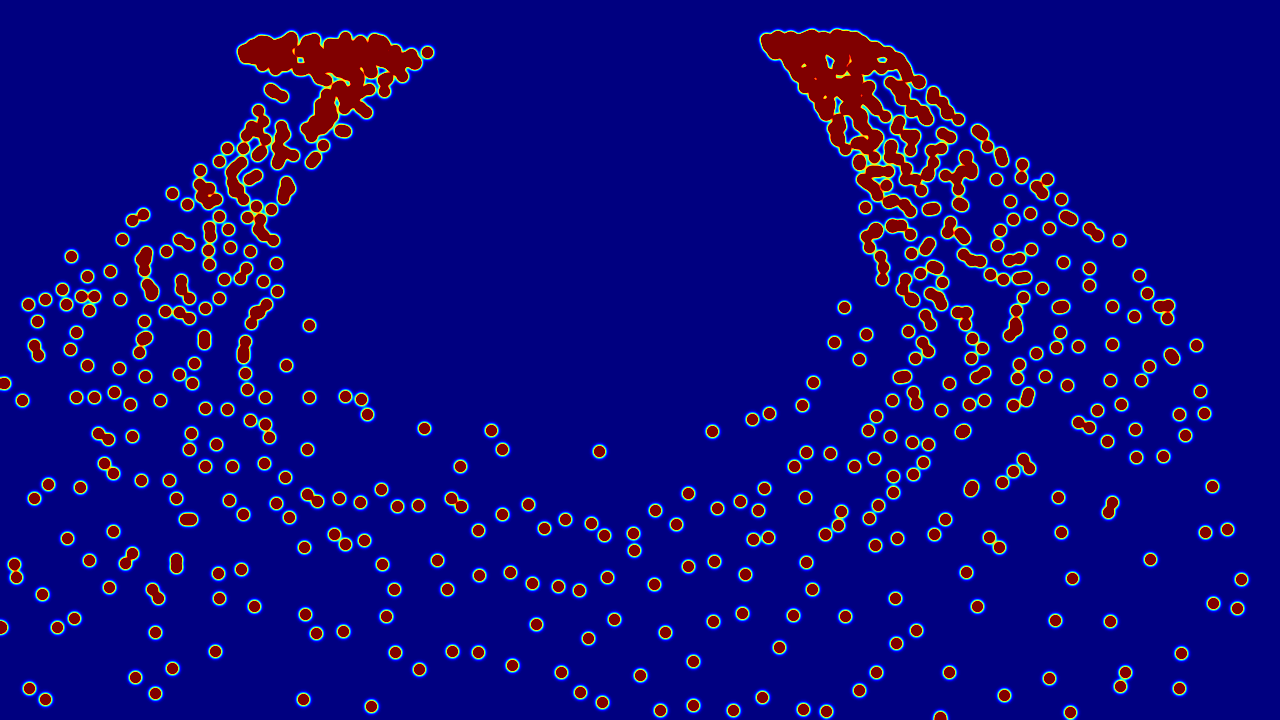}&
 \includegraphics[width=.245\linewidth]{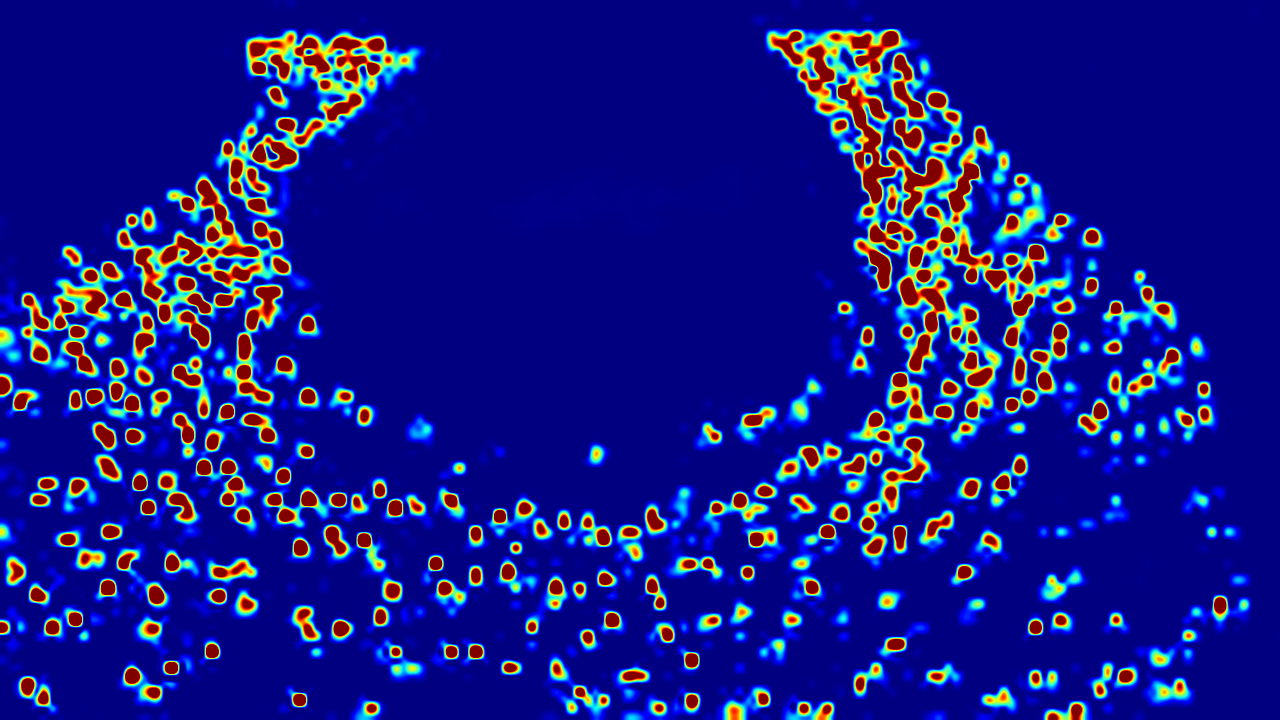}&
\includegraphics[width=.245\linewidth]{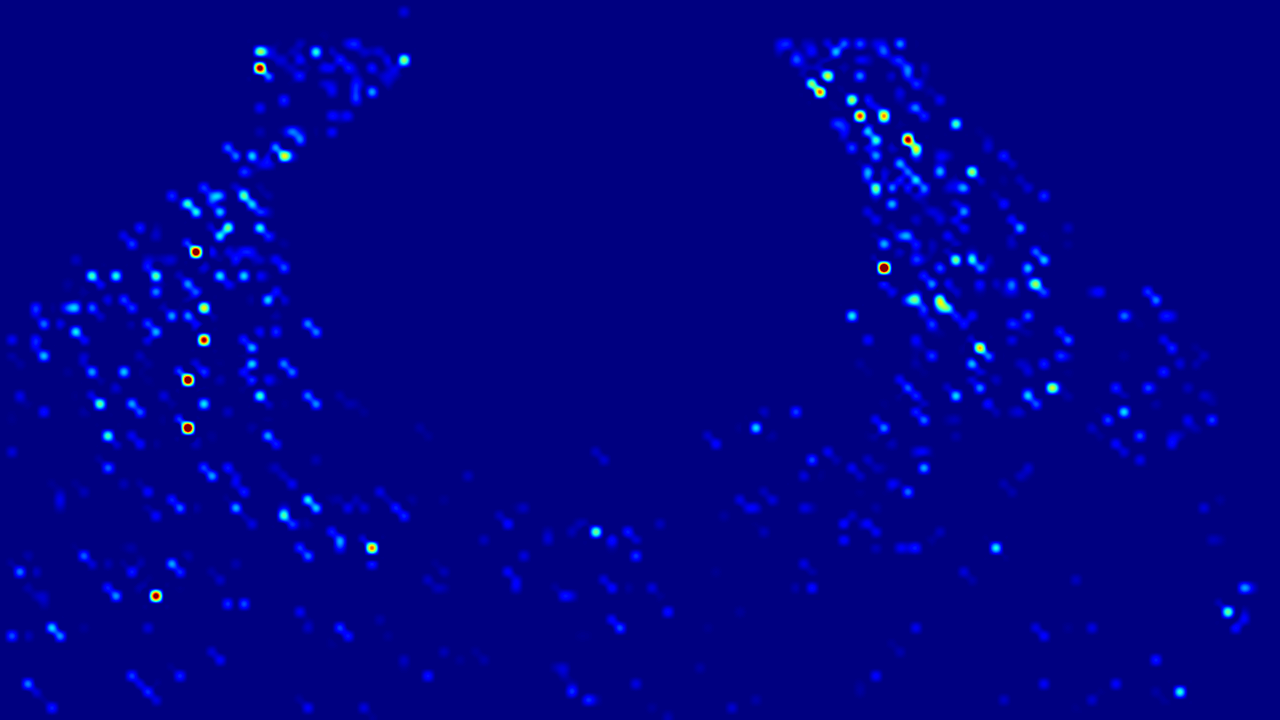}\\
\tiny{(a) original image }&
\tiny{(b) ground truth density map} &
\tiny{(c) estimated density map}&
\tiny{(d) flow direction $\nwarrow$}\\
\includegraphics[width=.245\linewidth]{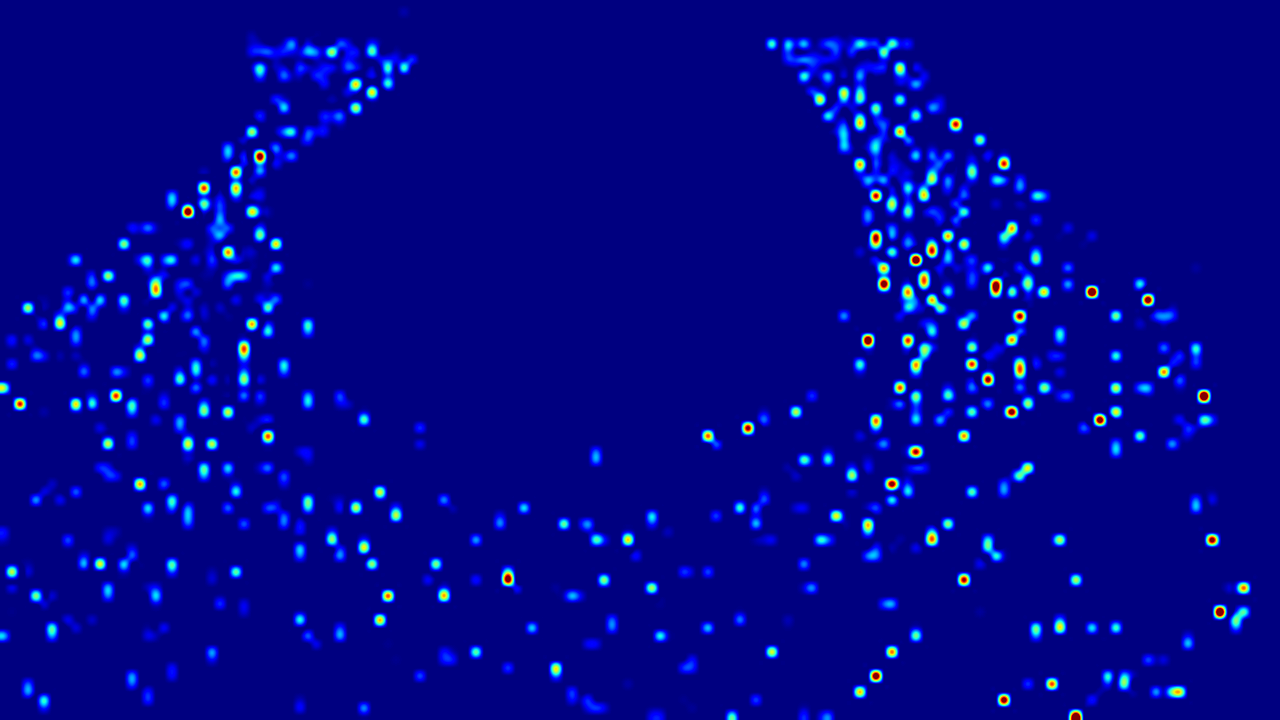}&
\includegraphics[width=.245\linewidth]{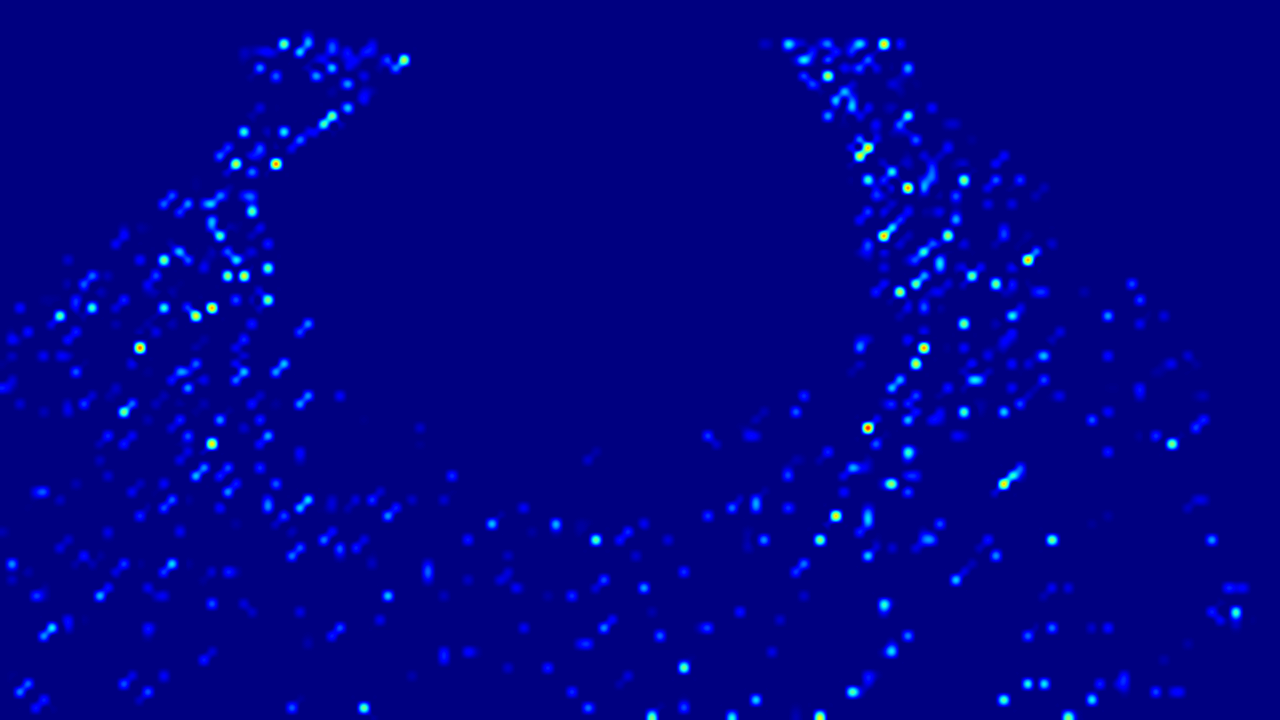}&
\includegraphics[width=.245\linewidth]{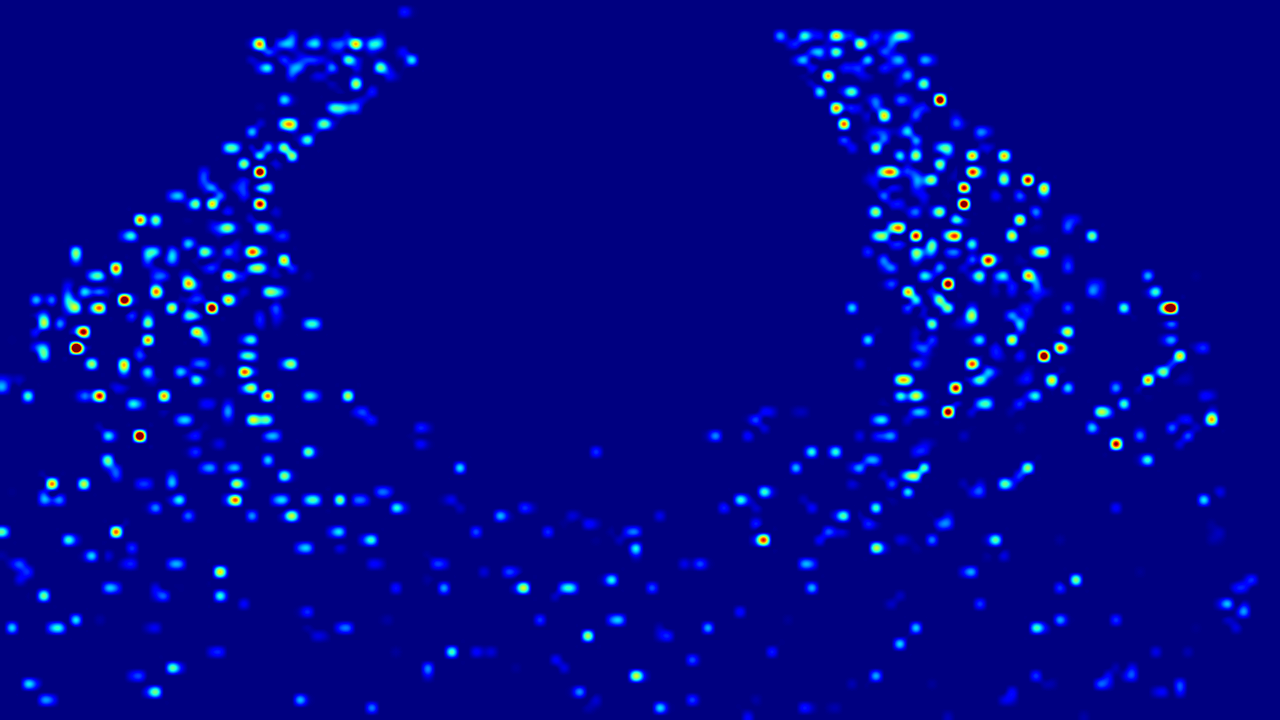}&
\includegraphics[width=.245\linewidth]{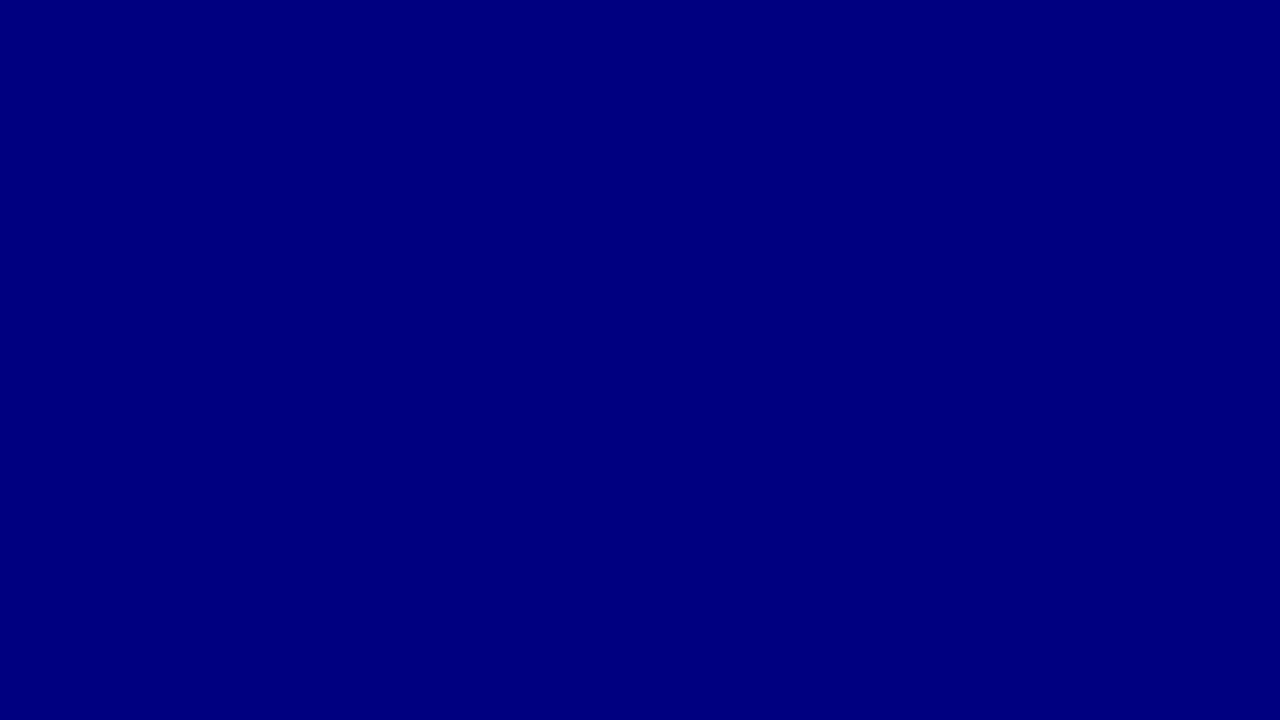}\\
\tiny{(e) flow direction $\uparrow$}&
\tiny{(f) flow direction $\nearrow$} &
\tiny{(g)	flow direction $\leftarrow$} & 
\tiny{(h) flow direction $\circ$}\\[1mm]
\includegraphics[width=.245\linewidth]{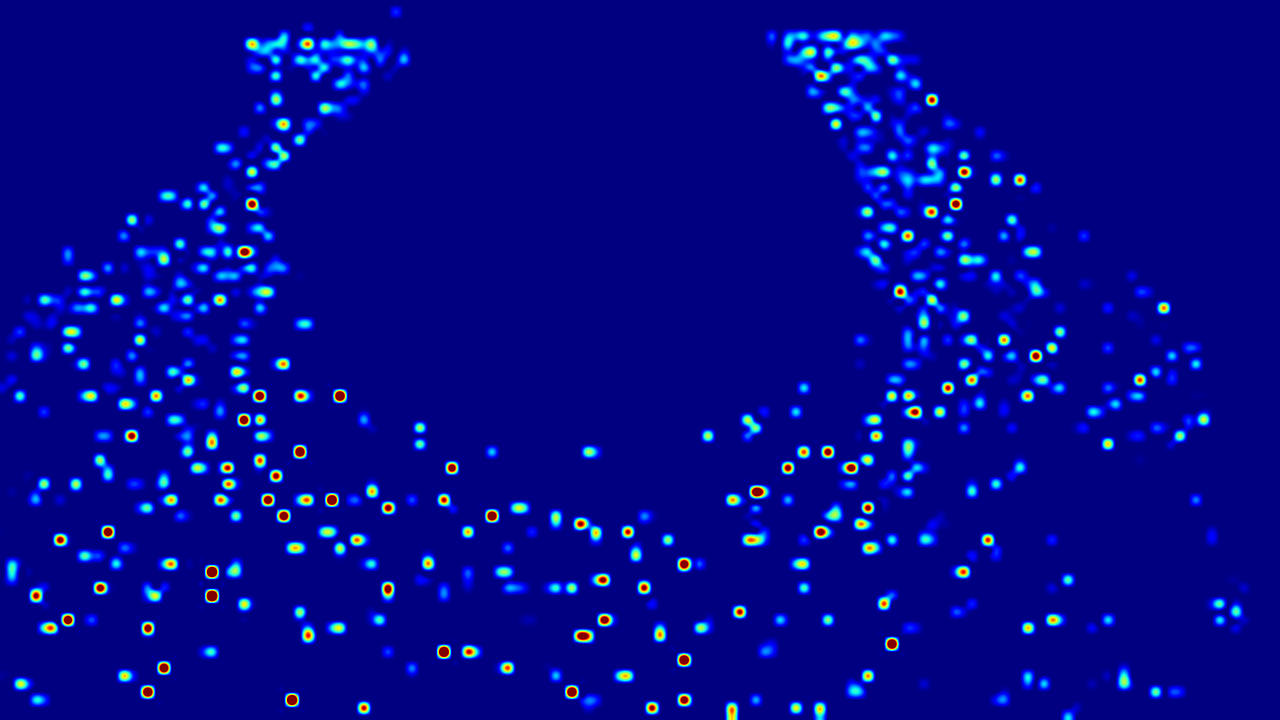}&
\includegraphics[width=.245\linewidth]{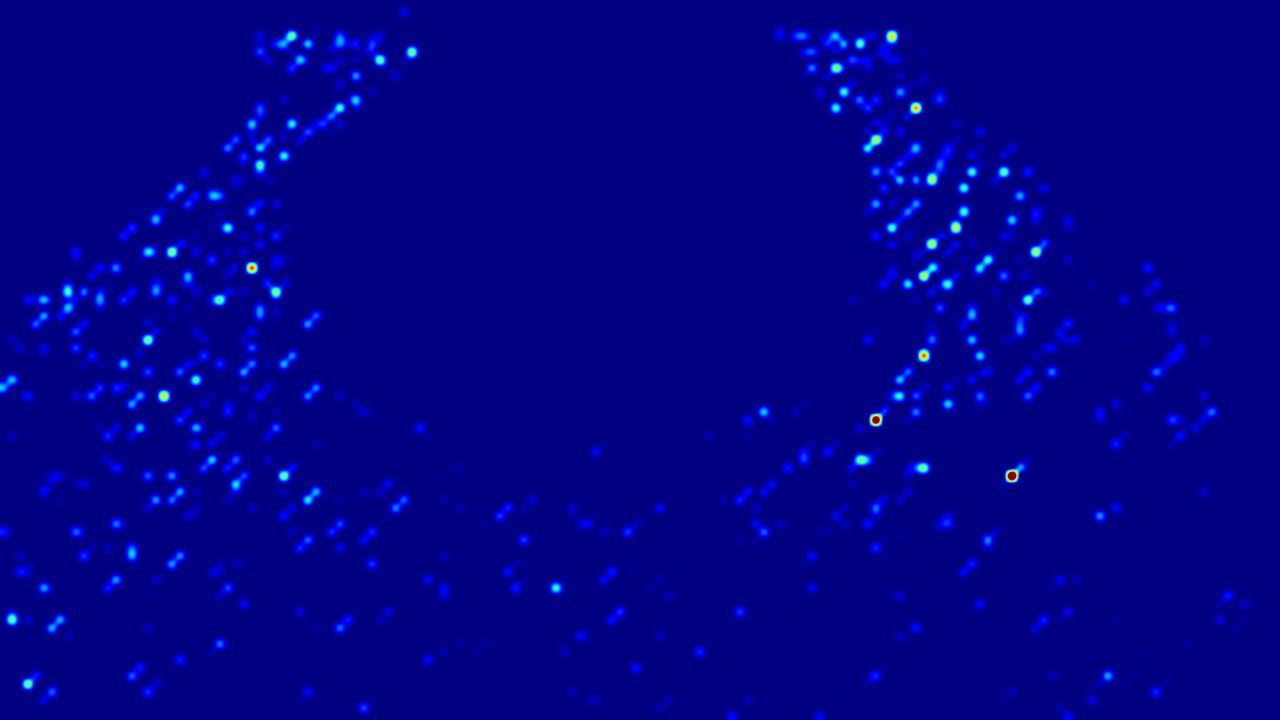}&
\includegraphics[width=.245\linewidth]{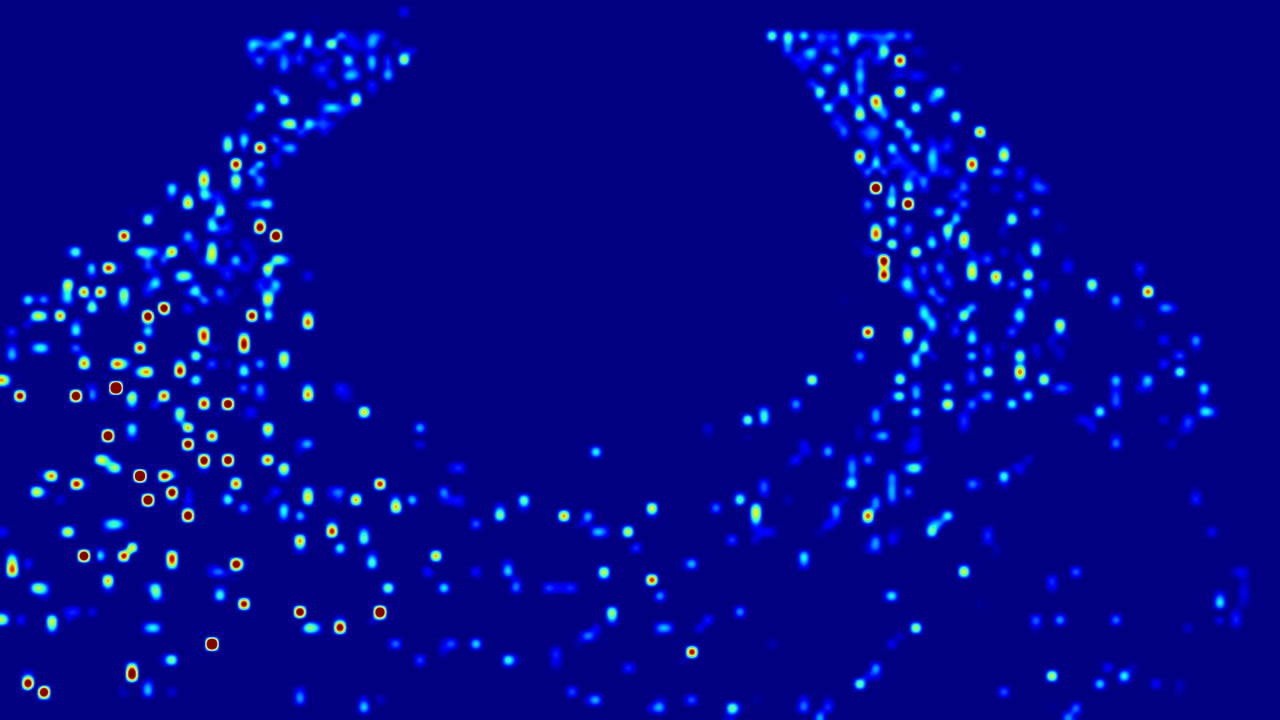}&
\includegraphics[width=.245\linewidth]{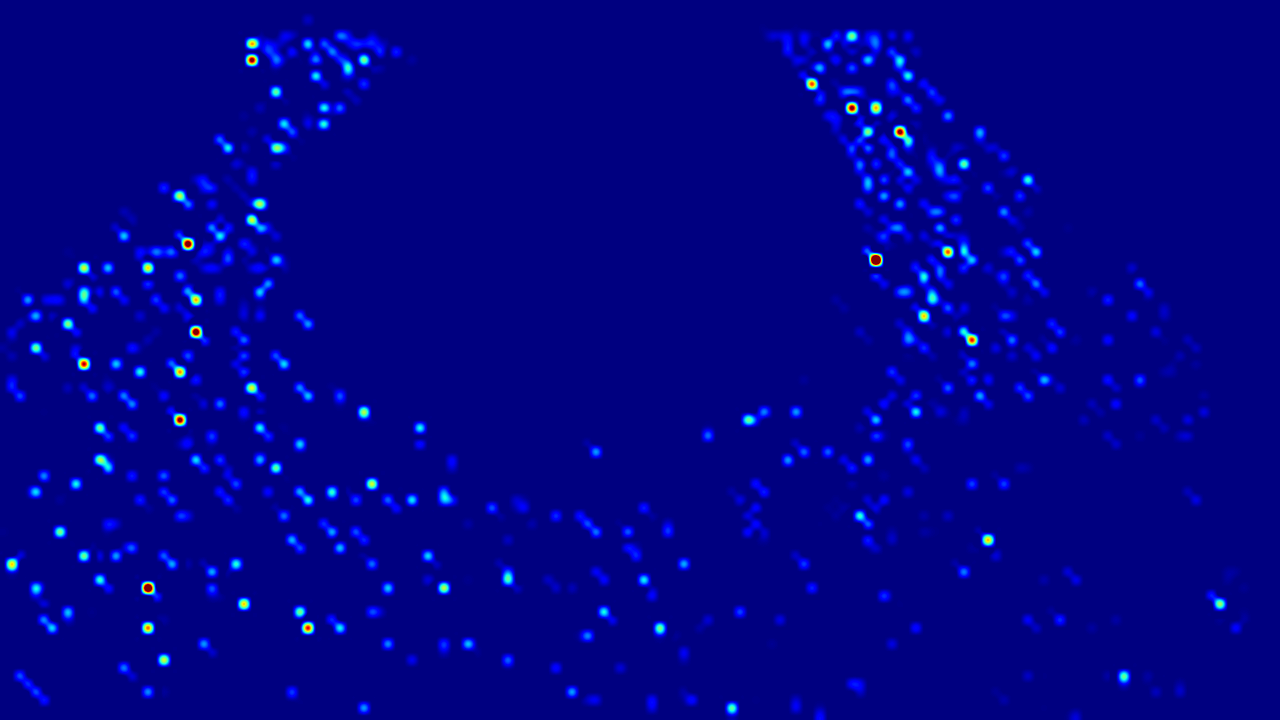}\\
\tiny{(i) flow direction $\rightarrow$}&
\tiny{(j) flow direction $\swarrow$} &
\tiny{(k) flow direction $\downarrow$} & 
\tiny{(l) flow direction $\searrow$}
\end{tabular}
\vspace{-3mm}
  \caption{ {\bf Density estimation in CrowdFlow.}  People are running counterclockwise. The estimated people density map is close to the ground-truth one. It was obtained by summing the flows towards the 9 neighbors of Fig.~\ref{fig:flow}~(b). They are denoted by the arrows and the circle. The latter corresponds to people not moving and is, correctly, empty. Note that the flow of people moving down is highest on the left of the building, moving right below the building, and moving up on the right of the building, which is also correct. Inevitably, there is also some noise in the estimated flow, some of which is attributable to body shaking while running.}
  \label{fig:crowdflowDensity}
  \end{figure*}

\parag{Synthetic Data.}
Fig.~\ref{fig:crowdflowDensity} depicts a qualitative result, and we report our quantitative results on the {\bf CrowdFlow} dataset in Table~\ref{tab:eva}~(a). \oursC{} outperforms the competing methods by a significant margin while \oursO{} delivers a further improvement. Using the ground-truth optical flow values in our $L_{all}$ loss term yields yet another performance improvement, that points to the fact that using better optical flow estimation than {\bf PWC-Net}~\cite{Sun18a} might help.


\begin{figure*}[htbp]
\centering
\begin{tabular}{cccc}
\includegraphics[width=.245\linewidth]{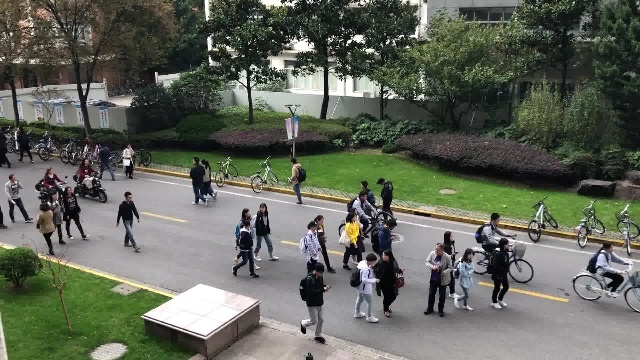}&
\includegraphics[width=.245\linewidth]{images/fdst/gt.jpg}&
\includegraphics[width=.245\linewidth]{images/fdst/pred.jpg}&
\includegraphics[width=.245\linewidth]{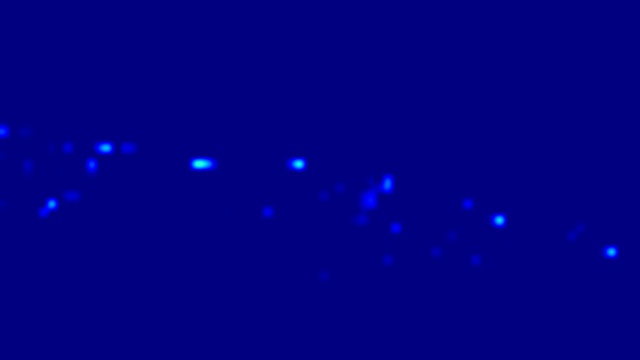}\\
 \tiny{(a) original image }&
 \tiny{(b) ground truth density map} &
 \tiny{(c) estimated density map}&
 \tiny{(d) flow direction $\nwarrow$}\\
 \includegraphics[width=.245\linewidth]{images/fdst/2.jpg}&
 \includegraphics[width=.245\linewidth]{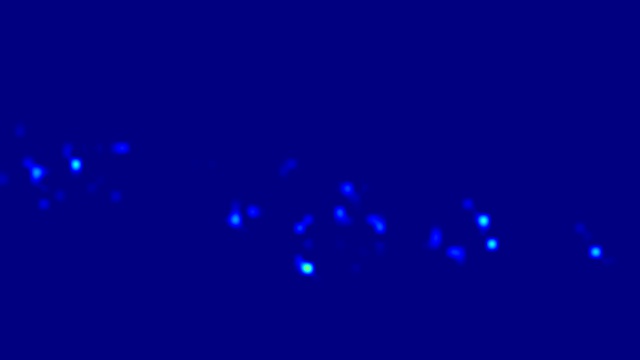}&
 \includegraphics[width=.245\linewidth]{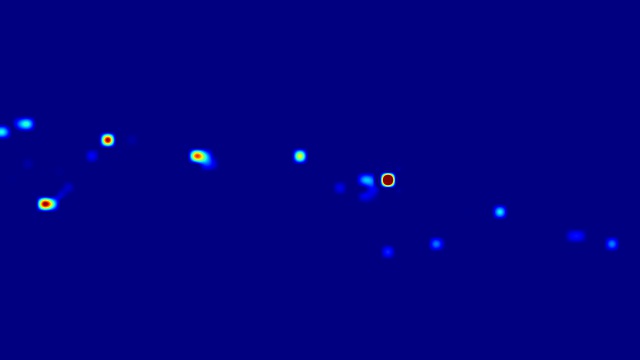}&
 \includegraphics[width=.245\linewidth]{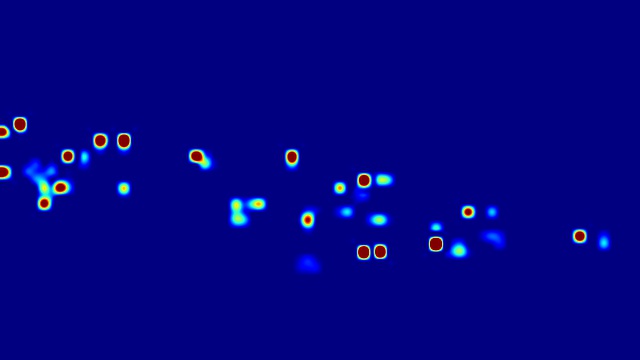}\\
 \tiny{(e) flow direction $\uparrow$}&
 \tiny{(f) flow direction $\nearrow$} &
 \tiny{(g)	flow direction $\leftarrow$} & 
 \tiny{(h) flow direction $\circ$}\\[1mm]
 \includegraphics[width=.245\linewidth]{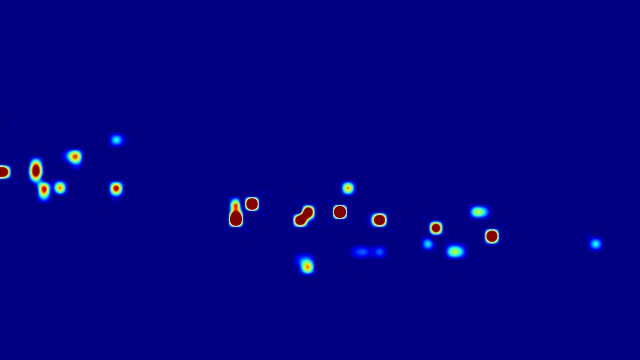}&
 \includegraphics[width=.245\linewidth]{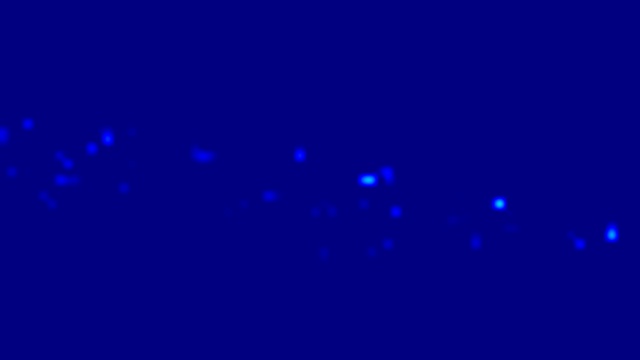}&
 \includegraphics[width=.245\linewidth]{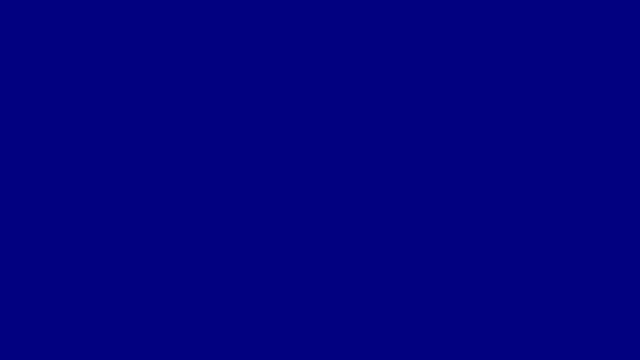}&
 \includegraphics[width=.245\linewidth]{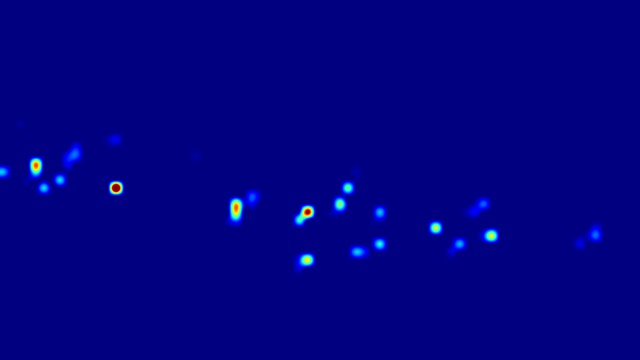}\\
 \tiny{(i) flow direction $\rightarrow$}&
 \tiny{(j) flow direction $\swarrow$} &
 \tiny{(k) flow direction $\downarrow$} & 
 \tiny{(l) flow direction $\searrow$}
\end{tabular}
\vspace{-3mm}
  \caption{ {\bf Density estimation in FDST.}  People mostly move from left to right. The estimated people density map is close to the ground-truth one. It was obtained by summing the flows towards the 9 neighbors of Fig.~\ref{fig:flow}~(b). They are denoted by the arrows and the circle. Strong flows occur in (g),(h), and (i), that is, moving left, moving right, or not having moved. Note that the latter does not mean that the people are static but only that they have not had time to change grid location between the two time instants.}
  \label{fig:fdstDensity}
  \end{figure*}

\parag{Real Data.} 
Fig.~\ref{fig:fdstDensity} depicts a qualitative result, and we report our quantitative results on the four real-world datasets in Tables~\ref{tab:eva}~(b), (c), (d) and (e). For {\bf FDST} and {\bf UCSD}, annotations in consecutive frames are available, which enabled us to pre-train the $\cF_o$ regressor of Eq.~\ref{eq:loss_optical}. By contrast, for {\bf Venice} and {\bf WorldExpo'10}, only a sparse subset of frames are annotated, and we therefore warp the crowd annotation using optical flow estimation from {\bf PWC-NET}~\cite{Sun18a}. We report results for both \oursC{} and \oursO{}. 

For {\bf FDST}, {\bf UCSD}, and {\bf Venice}, our approach again clearly outperforms the competing methods, with the optical flow constraint further boosting performance when applicable. For {\bf WorldExpo'10}, the ranking of the methods depends on the scene being used, but ours still performs best on average and on Scene3. In short, when the crowd is dense, our approach dominates the others. By contrast, when the crowd  becomes very sparse as in  Scene1 and Scene5, models that comprise a pool of different regressors, such as~\cite{Shi18}, gain an advantage. This points to a potential way to further improve our own method, that is, to also use a pool of regressors to estimate the people flows.

Recall that for {\bf FDST} and {\bf UCSD} all training frames are annotated whereas only a fraction are for  {\bf Venice}  and {\bf WorldExpo'10}, which demonstrates that our approach can handle a large number of unannotated frames. In the supplementary material, we re-run our training using only a fraction of the annotated frames in {\bf FDST} and {\bf UCSD} and demonstrate graceful performance degradation.


\begin{figure}[htbp]
\centering
\begin{tabular}{cc}
\includegraphics[width=.45\linewidth]{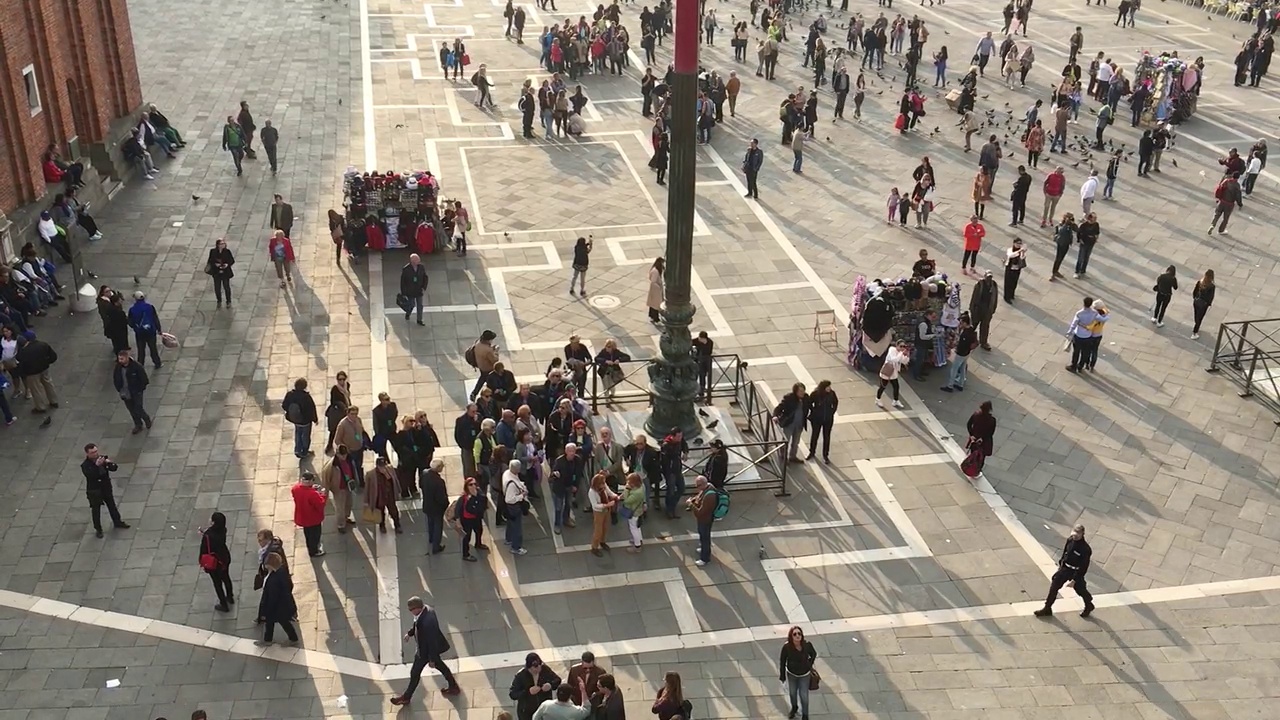}&
\includegraphics[width=.45\linewidth]{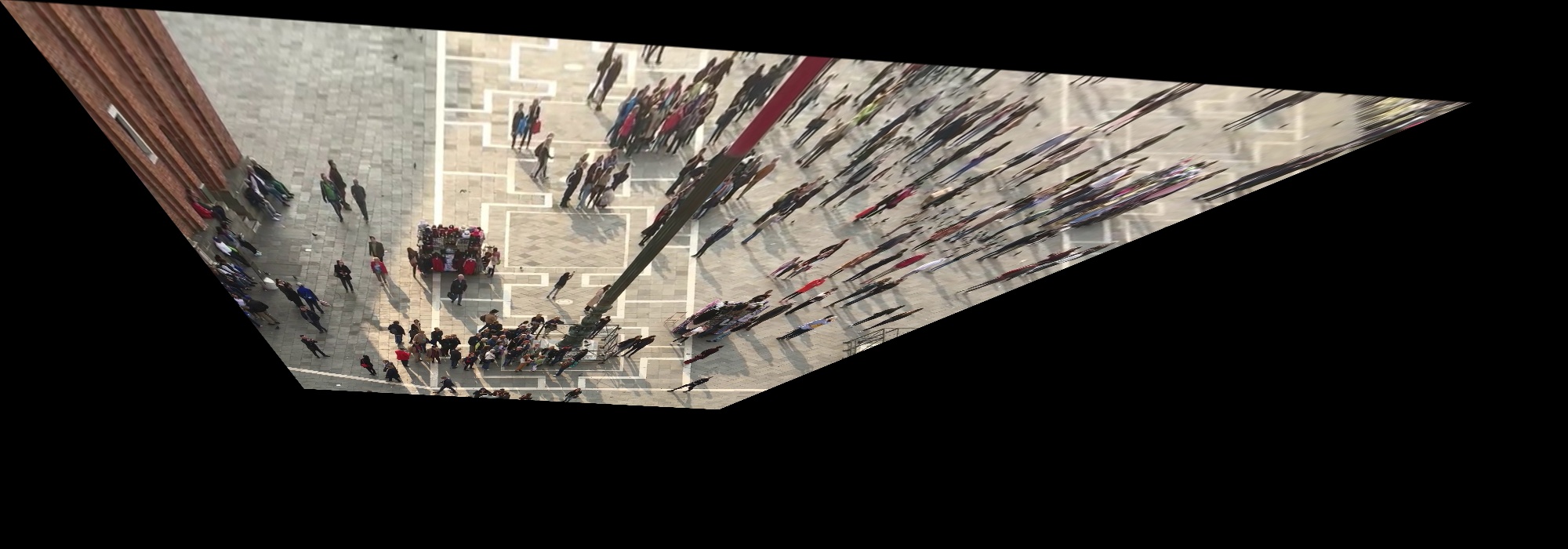}\\
\tiny{(a) image plane image }&
\tiny{(b) ground plane image} \\
\includegraphics[width=.45\linewidth]{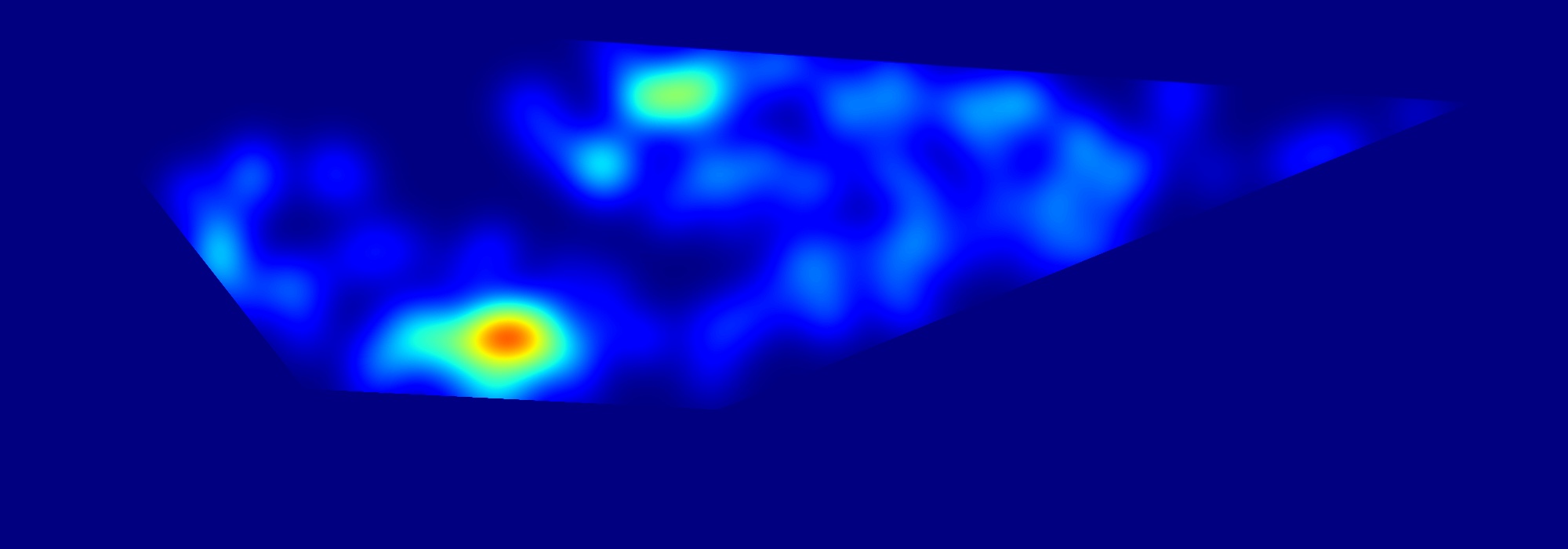}&
\includegraphics[width=.45\linewidth]{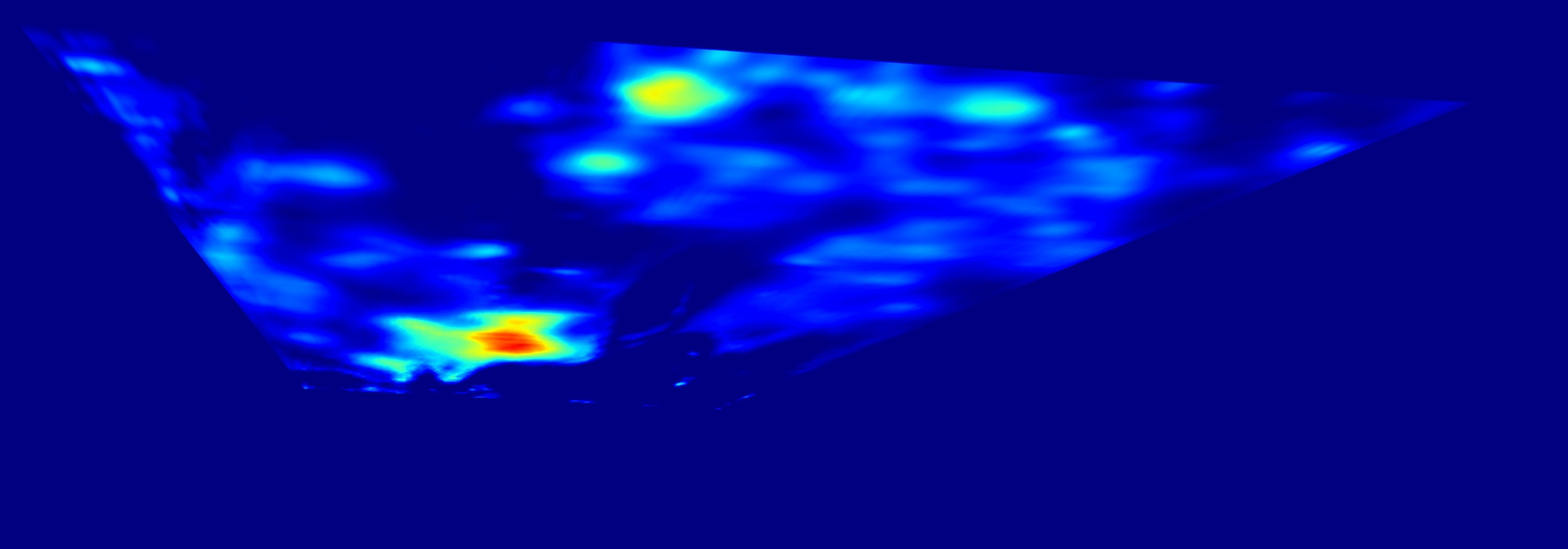}\\
 \tiny{(c) ground truth ground plane density map}&
 \tiny{(d) estimated ground plane density map}\\
\end{tabular}
\vspace{-3mm}
  \caption{ {\bf Ground plane density estimation in Venice.}  An image and its corresponding ground plane density map estimation.}
  \label{fig:ground}
  \end{figure}


\begin{figure*}[htbp]
\centering
\begin{tabular}{ccc}
 \includegraphics[width=.3\linewidth]{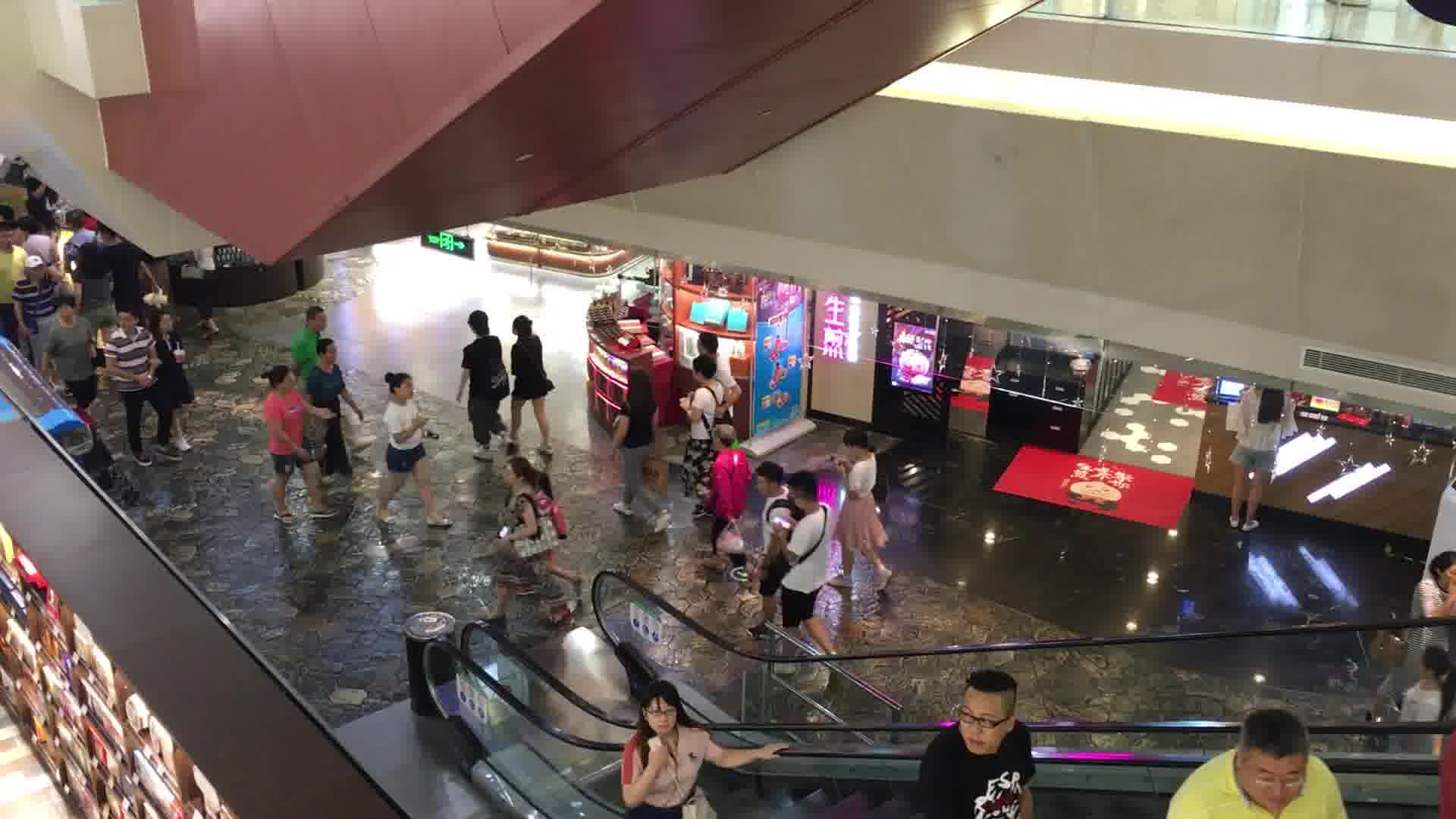}&
 \includegraphics[width=.3\linewidth]{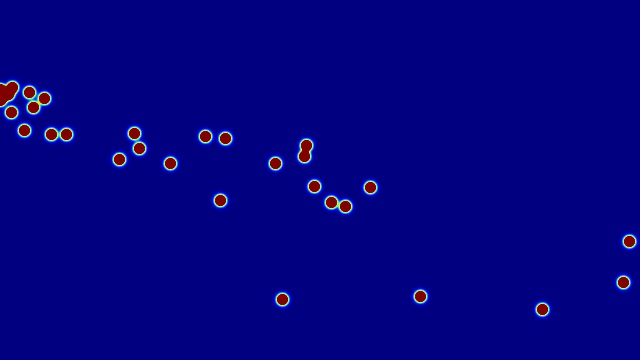}&
 \includegraphics[width=.3\linewidth]{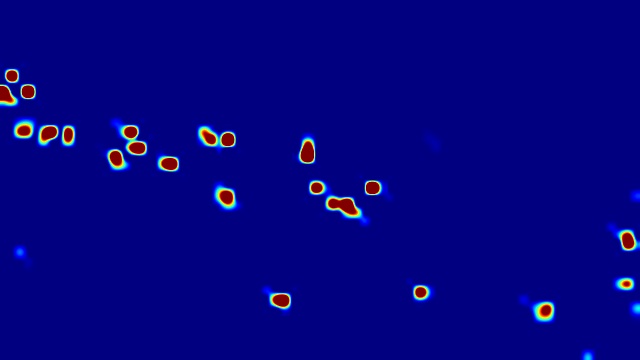}\\
 \footnotesize{(a) Input image }&
 \footnotesize{(b) Ground truth crowd density map} &
 \footnotesize{(b) Our prediction}
\end{tabular}
\vspace{-3mm}
  \caption{ {\bf Example crowd density map prediction with less annotation.} (a) Example test image from {\bf FDST}~\cite{Fang19a} dataset (b) Ground truth crowd density map (c) Inferred crowd density map. Note how similar our prediction is to the ground truth one even though only a $1/16$ patch of each image is annotated in the training dataset.}
    \label{fig:fdst}
\end{figure*}

\subsubsection{Working in the Ground Plane}

Until now, we have performed all the computations in image space, in large part so that we can compare our results to that of other recent algorithms that also work in image space. However, this neglects perspective effects as people densities per unit of image area are affected by where in the image the pixels are. To account for them, we can model them by working in the ground plane instead of the image plane, which we do in this section. 
    
Let $\bH^i$ be the homography from image $I^i$ to the corresponding ground plane. We define the ground-truth density as a sum of Gaussian kernels centered on peoples' heads on the ground plane. Because we now work in the physical world, we can use the same kernel size across the entire scene and across all scenes. A head annotation $P^i$, that is, a 2D image point expressed in projective coordinates, is mapped to $\bH^i P^i$ on the ground plane. Given a set $A^{i}=\{P^i_1,...,P^i_{c_i}\}$ of $c^i$ such annotations, we take the {\it ground plane density} $G^i$ at point $P$ expressed in ground plane coordinates to be
\begin{equation}
    G^i(P)= \sum_{j=1}^{c^i} \mathcal{N}(P| \bH^i P^i_j, \sigma) \; ,
    \label{eq:densityGround}
\end{equation}
where $\mathcal{N}(. | \mu,\sigma)$ is a 2D Gaussian kernel with mean $\mu$ and variance $\sigma$. Note the difference compared with image plane crowd density, which is defined at Eq.~\ref{eq:densityMap}. If we take our grid cells to be 30cm square and use a 30 FPS video, no one going slower than 9m/s, i.e., 32.5 km/h, can exit the neighborhood of its current location between two frames, which is more than enough for most humans. For faster animals, we would have to work with larger grid cells, more extended neighborhoods, or a higher frame rate.

Since {\bf Venice} is the only publicly available video-based single-view crowd counting dataset containing accurate camera pose information, it is the one we used to evaluate this approach. In the supplementary material, we also evaluate results in the ground plane on several multi-view crowd counting datasets. The ground plane regressor architecture is the same as before, with an additional Spatial Transformer Networks~\cite{Jaderberg15} to map the output to the ground plane. The results are denoted by \oursGP{} in Table~\ref{tab:eva}(c) and show  a marked improvement over \oursC{} that operates strictly in the image plane. Fig.~\ref{fig:ground} depicts corresponding density estimates in the image and ground planes.

\subsubsection{Ablation Study}

We know examine the individual components of our fully-supervised approach and show that each one contributes to these results. 


\begin{table}
  \centering
  \scalebox{0.6}{
\begin{tabular}{lccccccccc}
  \toprule
  &People  & Cycle  & Optical  &\multicolumn{2}{c}{{\bf CrowdFlow}} &\multicolumn{2}{c}{{\bf UCSD}} &\multicolumn{2}{c}{{\bf FDST}}\\
  Model  & Flow& Consistency& Flow & $MAE$ & $RMSE$  & $MAE$ & $RMSE$  & $MAE$ & $RMSE$ \\
  \midrule
  \baseline{} & & & & 124.3 & 160.2   & 0.98 & 1.26 & 2.44 & 2.96\\
  \twos{} & & & & 125.7 & 164.1  & 1.02 & 1.40 & 2.48 & 3.10\\
  \ave{} & & & & 128.9 & 174.6  & 1.01 & 1.31  & 2.52 & 3.14\\
  \weak{}~\cite{Liu19b} & & & & 121.2 & 155.7 & 0.96 & 1.30  & 2.42 & 2.91 \\
  \oursF{} & \checkmark  & & & 113.3 & 140.3  &0.94  &1.21 &2.31  &2.85\\
  \oursC{} & \checkmark & \checkmark & & 97.8 & 112.1   & 0.86 & 1.13 & 2.17 & 2.62\\
  \oursO{}& \checkmark& \checkmark&\checkmark &\textbf{96.3} & \textbf{111.6}  & \textbf{0.81} & \textbf{1.07} & \textbf{2.10} & \textbf{2.46}\\
  \bottomrule 
  \end{tabular}
  }
  \caption{{\bf People flow vs people densities.} The tick marks indicate what subset of the consistency constraints each method uses.}
    \label{tab:ablation1}
\end{table}

\begin{table}
  \centering
  \scalebox{0.6}{
\begin{tabular}{lccccccccc}
  \toprule
  &People  & Cycle  & Optical  &\multicolumn{2}{c}{{\bf CrowdFlow}} &\multicolumn{2}{c}{{\bf UCSD}} &\multicolumn{2}{c}{{\bf FDST}}\\
  Model  & Flow& Consistency& Flow & $MAE$ & $RMSE$  & $MAE$ & $RMSE$  & $MAE$ & $RMSE$ \\
  \midrule
   \oursI{} & \checkmark&\checkmark & \checkmark& 97.5 & 110.7   & 0.85 & 1.21 & 2.15 & 2.74\\
   \oursO{}& \checkmark& \checkmark&\checkmark &\textbf{96.3} & \textbf{111.6}  & \textbf{0.81} & \textbf{1.07} & \textbf{2.10} & \textbf{2.46}\\
   \bottomrule 
  \end{tabular}
  }
  \caption{{\bf Training the optical flow regressor}}
    \label{tab:ablation2}
\end{table}

\begin{table}
  \centering
  \scalebox{0.6}{
\begin{tabular}{lccccccccc}
  \toprule
  &People  & Cycle  & Optical  &\multicolumn{2}{c}{{\bf CrowdFlow}} &\multicolumn{2}{c}{{\bf UCSD}} &\multicolumn{2}{c}{{\bf FDST}}\\
  Model  & Flow& Consistency& Flow & $MAE$ & $RMSE$  & $MAE$ & $RMSE$  & $MAE$ & $RMSE$ \\
  \midrule
  \oursCF{} & \checkmark & \checkmark & & 98.0 & 112.6   & 0.87 & 1.19 & 2.19 & 2.65\\
  \oursCB{} & \checkmark & \checkmark & & 98.1 & 112.4   & 0.88 & 1.14 & 2.18 & 2.63\\
  \oursS{} & \checkmark & \checkmark & & {\bf 97.7} & 112.4   & {\bf 0.86} & 1.15 & 2.18 & {\bf 2.59}\\
   \oursC{} & \checkmark & \checkmark & & 97.8 & {\bf 112.1}   & {\bf 0.86} & {\bf 1.13} & {\bf 2.17} & 2.62\\
   \bottomrule 
  \end{tabular}
  }
  \caption{{\bf Using the spatial loss term and reversing the flows}}
    \label{tab:ablation3}
\end{table}

\parag{People Flows vs People Densities.}
To confirm that the good performance we report really is attributable to our regressing flows instead of densities, we performed the following set of experiments. Recall from Section~3.2, that we use the CAN~\cite{Liu19a} architecture to regress the flows. Instead, we can use it to directly regress the densities, as in the original CAN paper. We will refer to this approach as \baseline{}. In~\cite{Liu19b}, it was suggested that people conservation constraints could be added by incorporating a loss term that enforces the conservation constraints of Eq.~\ref{eq:conservation} that are weaker than those of Eq.~\ref{eq:flow}, that is, those we use in this paper. We will refer to this approach relying on weaker constraints while still using the CAN backbone as \weak{}. As \oursC{}, it takes two consecutive images as input. For the sake of completeness, we also implemented a simplified approach, \twos{}, that takes the same two images as input  and directly regresses the densities. To show that regressing flows is more effective than simply smoothing the densities, we implement \ave{} that takes three images as input, uses CAN to independently compute three density maps, and then averages them. Finally, to highlight the importance of the forward-backward constraints of Eq.~\ref{eq:flowConstraints}, we also tested a simplified version of our approach in which we drop them and that we refer to as \oursF{}.

We compare the performance of these five approaches on {\bf CrowdFlow}, {\bf FDST}, and  {\bf UCSD} in Table~\ref{tab:ablation1}. Both \twos{} and \ave{} do worse than \baseline{}, which confirms that temporal averaging of the densities is not the right thing to do. As reported in~\cite{Liu19b}, \weak{} delivers  a small improvement.  As expected \oursF{} improves on \twos{} in all three datasets, with further performance increase for \oursC{} and \oursO{}. This confirms that using people flows instead of densities is a win and that the additional constraints we impose all make positive contributions.

\parag{Training the Optical Flow Regressor.}
As explained in Section~\ref{sec:flow}, we use optical flow to regularize the people flow estimates. To this end, we need to train the regressor $\cF_o$ of Eq.~\ref{eq:loss_optical} that associates to consecutive density images an optical flow estimate that can be compared to that produced by a state-of-the-art optical flow estimator. In our implementation, $\cF_o$ takes as input the density images but  {\it not} the original images, our intuition being that if it did, it could predict the correct optical flows even if the density estimates were wrong, which would defeat its purpose. To confirm this, we implemented a version called \oursI{} in which $\cF_o$ takes both the original images and crowd density maps as input. As can be seen in Table~\ref{tab:ablation2}, the results are less good.

\parag{Using the Spatial Loss Term.}
Our active learning approach of Section~\ref{sec:AL} relies on the spatial loss term $L_{spatial}$ of Eq.~\ref{eq:spatialLoss}, which we do not normally use in the fully-supervised case, essentially because minimizing it imposes constraints that are weaker than those than the flow-consistency constraints of Eq.~\ref{eq:flowConstraints} impose. To check the validity of this choice, we implemented a variant of our approach that includes this additional loss term and that we refer to as \oursS{}. As can be in seen in Table~\ref{tab:ablation3}, it performs very comparably to \oursC{}, as could be expected.

\parag{Forward vs Backward Flows.}
In our approach we compute both forward flows of the form $f^{t-1,t}_{i,j}$ and backward flows of the  form $f^{t,t-1}_{j,i}$ and we can sum either to obtain the people densities. Let \oursCF{} and \oursCB{} be versions of our approach that does either, whereas  \oursC{} averages the two values, which provides a slight boost as can be seen in Table~\ref{tab:ablation3}. 

\parag{Distance between Annotated Frames.}
We refer to the number of frames between annotated frames in the training set as $V$.  For {\bf CrowdFlow}, {\bf FDST} and {\bf UCSD}, $V=1$ because all frames are annotated. For {\bf Venice} and {\bf WorldExpo'10}, annotations are available for every $60$ and $255$ frames. Hence, $V=60$ and $V=255$, respectively. We re-run our training on  {\bf CrowdFlow}, {\bf FDST} and {\bf UCSD} for $V=2$ and $V=5$ and report the results in Table~\ref{tab:v_value}. Even using $V=1$ enforces stronger constraints, $V=2$ delivers almost the same performance. For $V=5$, there is a performance decrease but it is relatively small considering that we are now using only one fifth of the annotations.


\begin{table}[tp]
  \centering
\scalebox{1.0}{
\begin{tabular}{lcccccc}
  \toprule
  &\multicolumn{2}{c}{{\bf CrowdFlow}} &\multicolumn{2}{c}{{\bf FDST}} &\multicolumn{2}{c}{{\bf UCSD}}\\
   $V$ & $MAE$ & $RMSE$ & $MAE$ & $RMSE$ & $MAE$ & $RMSE$ \\
  \midrule
  $V = 1$  & {\bf 97.8} & 112.1 & {\bf 2.17} & {\bf 2.62} & {\bf 0.86} & 1.13 \\
  $V = 2$  & 98.1 &  {\bf 111.9} & 2.20 & 2.66 & {\bf 0.86} & {\bf 1.12} \\ 
  $V = 5$  & 106.2 & 124.2 & 2.63 & 3.36 & 0.90 & 1.28 \\
  \bottomrule
  \end{tabular}
}
  \caption{{\bf Ablation study of annotation interval value on {\bf CrowdFlow}, {\bf FDST} and {\bf UCSD}.}}
  \label{tab:v_value}

\end{table}

\subsection{Active Learning with Self-Supervision}
\label{sec:active}


\begin{figure*}
\centering
\includegraphics[width=1.0\linewidth]{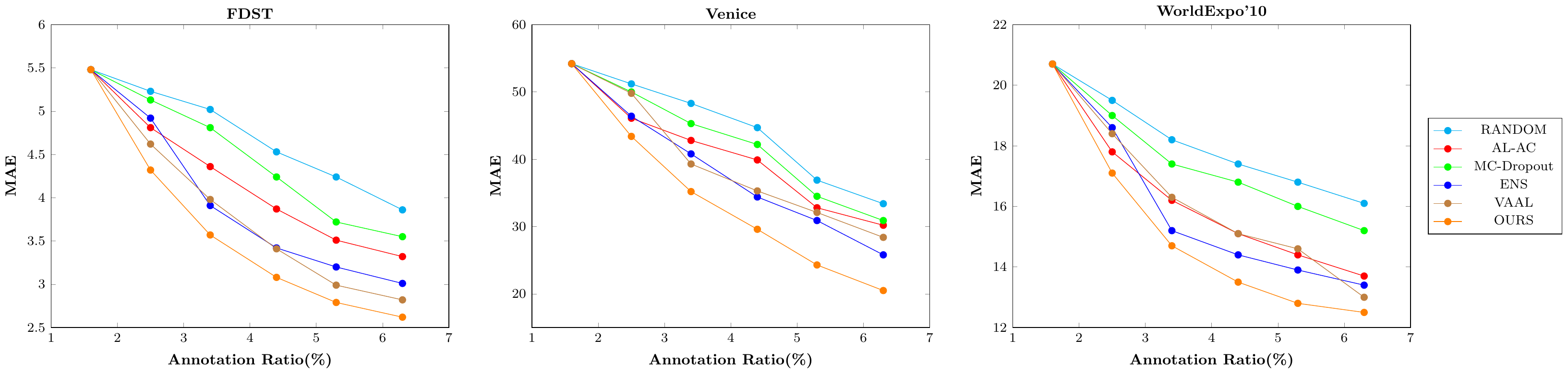}
  \caption{{\bf Comparing against other AL approaches.}  We plot the $MAE$ obtained using different active learning algorithms as a function of the annotation ratio. All models were initially trained with 25\% randomly selected images of which only $1/16$ of the area was annotated.  At each active learning iteration, another 15\% of the images were selected either randomly or actively and another $1/16$th annotated. All the models are trained using the same loss function, the only difference being how the patches are selected.  Our AL approach consistently outperform others in all the datasets.}
  \label{fig:eva_al}
  \end{figure*}


\begin{figure*}
\centering
\includegraphics[width=1.0\linewidth]{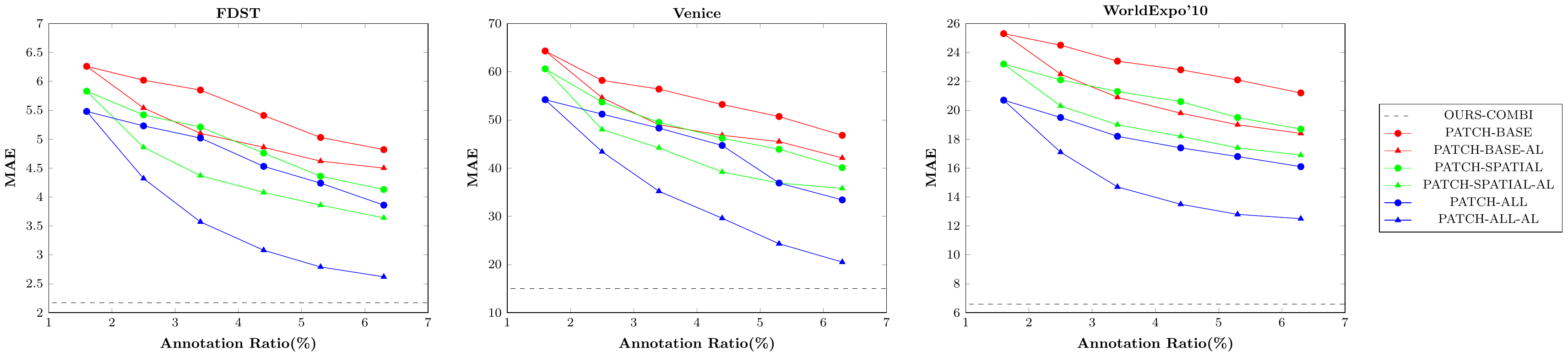}
  \caption{ {\bf Ablation study of our AL approach.}  We plot the $MAE$ obtained using different versions of our AL strategy as a function of the annotation ratio. As expected, our complete approach does best.}
  \label{fig:ablation_al}
  \end{figure*}

Recall from Section~\ref{sec:results} that \oursC{} denotes our full approach when taking a single image as input, that is, without exploiting temporal consistency. Here, we combine it with the active learning strategies for Section~\ref{sec:AL}.

\subsubsection{Comparing against Recent Techniques}

Here, we compare our patch selection strategy against other AL approaches in the same setting.
\begin{itemize}
    \item AL-AC~\cite{Zhao20a}: It is a recent approach to active crowd counting, it actively choose the unlabeled images with high dissimilarity in crowd density distribution compared with the labeled one. Besides, a discriminator classifier is also added to distinguish if the sample is labeled or not.
    
    \item MC-Dropout~\cite{Gal16}: It measures the uncertainty by sampling from the average output of multiple forward passes with random dropout masks. Samples with high uncertainty are selected for training in next iteration.
    
    \item ENS~\cite{Beluch18}: It is an ensemble-based approach which measures the uncertainty by sampling from the average output of multiple forward passes of different models trained with different initialization . Same as MC-Dropout, samples with high uncertainty are selected for training in next iteration.
    
    \item VAAL~\cite{Sinha19}: It learns a latent space using a variational auto encoder (VAE) and an adversarial network trained to discriminate between unlabeled and labeled data. The samples predicted to be unlabeled with high probability are chosen to annotate in next iteration.
    
\end{itemize}

We extend the above approaches in the same setting as ours with the same crowd density regressors. All models are trained using the same loss function $L_{overall}$ of Eq.~\ref{eq:overall}. The only difference is how we select the patches to annotate. We evaluate the various approaches on {\bf FDST}, {\bf Venice} and {\bf WorldExpo'10}. As can be seen in  Fig.~\ref{fig:eva_al}, our approach consistently outperforms the others.

\subsubsection{Ablation Study}

We now turn to the individual components of our active-learning scheme and implemented the following variants to gauge their impact:

\begin{itemize}

 \item \patchB{}. The model is trained using a single patch per image by only minimizing the supervised loss function $L_{combi}$ of Eq.~\ref{eq:loss1} and randomly selecting the patch to annotate. 
 
 \item  \patchBA{}. The model is trained using the same loss as \patchB{} but we actively select the patch to annotate using the measure of consistency violation of Eq.~\ref{eq:error}. 
 
 \item \patchS{}. The model is trained using the combined loss function including $L_{combi}$ and $L_{spatial}$ of Eq.~\ref{eq:loss1} and Eq.~\ref{eq:spatialLoss}; the patch is selected randomly. 
 
 \item  \patchSA{}. The model is trained using the same loss as \patchS{} but we actively select the patch to annotate using the measure of consistency violation of Eq.~\ref{eq:error}. 

 \item \patchA{}. The model is trained with the complete loss function $L_{overall}$ of Eq.~\ref{eq:overall};  the patch to annotate is selected randomly. 
 
 \item  \patchAA{}. The model is trained using the same loss as \patchS{} and we actively select the patch to annotate using the measure of consistency violation of Eq.~\ref{eq:error}. 

\end{itemize}

For all models, we start by randomly selecting 25\% of the training images, each of which is split into $4 \times 4$ patches, only one of which is annotated. Therefore, the starting annotation rate is $25\% / 16 = 1.5625 \%$. During each active learning iteration, another 15\% of the training images are selected, and we also annotate one patch of each image. After 5 iterations, only 6.25\% of the training patches have been selected, and we measured the ratio of annotated people to be around 5.7\%. Fig.~\ref{fig:ablation_al} depicts the $MAE$ on {\bf FDST}, {\bf Venice} and {\bf WorldExpo'10}. Note that both our loss terms and the AL algorithm consistently improve the performance with the largest boost coming from the active patch selection strategy. Furthermore, as can be seen by comparing these results with those in Tables~\ref{tab:eva}~(b),~(c) and~(e), even though \patchAA{} only uses 6.25\% of the annotations, it outperforms several SOTA models trained with full supervision. Fig.~\ref{fig:fdst} depicts an example density map inferred by \patchAA{}. Please refer to the supplementary material for an analysis of the influence of hyper-parameters choices.


\section{Conclusion}
We have shown that implementing a crowd counting algorithm in terms of estimating the people flows and then summing them to obtain people densities is more effective than attempting to directly estimate the densities. This is because it allows us to impose conservation constraints that make the estimates more robust. When optical flow data can be obtained, it also enables us to exploit the correlation between optical flow and people flow to further improve the results. Furthermore, we have demonstrated that spatial and temporal people conservation can be exploited to train a deep crowd counting model in an active learning fashion, achieving competitive performance with much fewer annotations.

In this paper, while we have mostly performed the computations in image space, in large part so that we could compare our results to that of other recent algorithms that also work in image space, we have also shown that modeling the people flows in the ground plane yields even better performance. A promising application is to use drones for people counting because their internal sensors can be directly used to provide the camera registration parameters necessary to compute the homographies between the camera and the ground plane. In this scenario, the drone sensors also provide a motion estimate, which can be used to correct the optical flow measurements and therefore exploit the information they provide as effectively as if the camera was static.

\ifCLASSOPTIONcompsoc
  \section*{Acknowledgments}
\else
  \section*{Acknowledgment}
\fi

This work was supported in part by the Swiss National Science Foundation.

\bibliographystyle{IEEEtran}
\bibliography{string,vision,learning}
\vspace{-1.3cm}
\begin{small}
\def \bioSpacing {-1.7cm}
\begin{IEEEbiography}[{\includegraphics[width=1in,height=1.25in,clip,keepaspectratio]{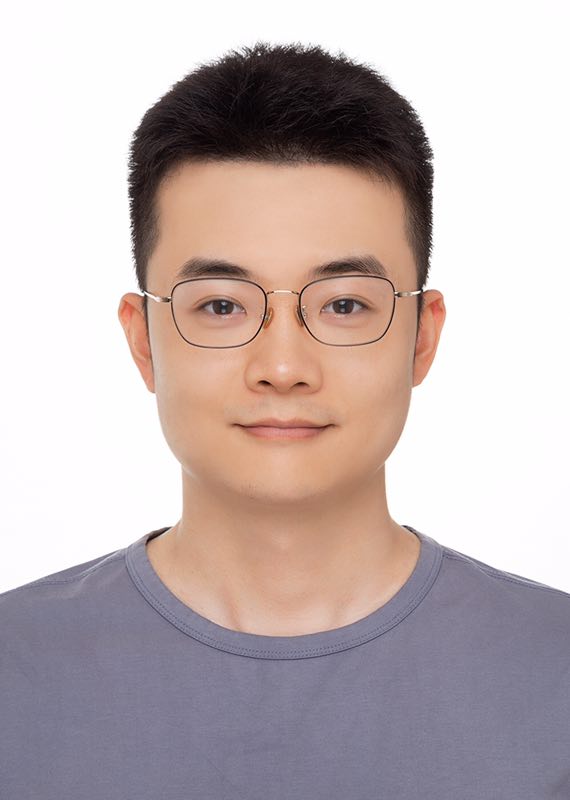}}]{Weizhe Liu}
	is a Ph.D. candidate and pursuing his Ph.D. degree in Computer Vision Laboratory, EPFL. Previously, he obtained his M.S. degree from EPFL in 2017 and the B.S. degree in University of Electronic Science and Technology of China in 2014. His research interests lie in the field of computer vision and machine learning.
\end{IEEEbiography}
\vspace{\bioSpacing}
\begin{IEEEbiography}[{\includegraphics[width=1in,height=1.25in,clip,keepaspectratio]{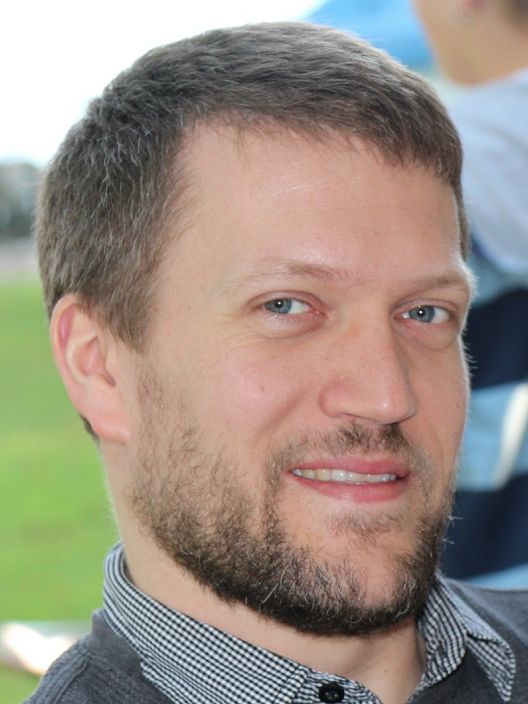}}]{Mathieu Salzmann}
	is a Senior Researcher at EPFL and an Artificial Intelligence Engineer at ClearSpace. Previously, he was a Senior Researcher and Research Leader in NICTA's computer vision research group, a Research Assistant Professor at TTI-Chicago, and a postdoctoral fellow at ICSI and EECS at UC Berkeley. He obtained his PhD in Jan. 2009 from EPFL. His research interests lie at the intersection of machine learning and computer vision.
\end{IEEEbiography}
\vspace{\bioSpacing}
\begin{IEEEbiography}[{\includegraphics[width=1in,height=1.25in,clip,keepaspectratio]{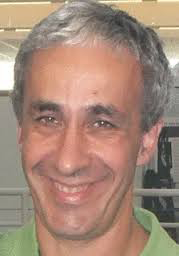}}]{Pascal Fua}
	is a Professor of Computer Science at EPFL, Switzerland. His research interests include shape and motion reconstruction from images, analysis of microscopy images, and Augmented Reality. He is an IEEE Fellow and has been an Associate Editor of the IEEE journal Transactions for Pattern Analysis and Machine Intelligence.
\end{IEEEbiography}
\end{small}

\end{document}